\documentclass[10pt,twocolumn,letterpaper]{article}

\usepackage{cvpr}
\usepackage{times}
\usepackage{epsfig}
\usepackage{graphicx}
\usepackage{amsmath}
\usepackage{amssymb}
\usepackage{epstopdf}
\usepackage{subfigure}
\usepackage{url}
\usepackage[pagebackref=true,breaklinks=true,letterpaper=true,colorlinks,bookmarks=false]{hyperref}

\cvprfinalcopy % *** Uncomment this line for the final submission

\def\cvprPaperID{1489} % *** Enter the CVPR Paper ID here

\ifcvprfinal\fi
\begin{document}
%%%%%%%%% TITLE
\title{MSR-net:Low-light Image Enhancement Using Deep Convolutional Network}

\author{Liang Shen$\thanks{Authors contributed equally.}$ , Zihan Yue$^*$, Fan Feng, Quan Chen, Shihao Liu, and Jie Ma\\
Institute of Image Recognition and Artificial Intelligence\\
Huazhong University of Science and Technology, Wuhan, China\\
}

\maketitle
\thispagestyle{empty}
\pagenumbering{arabic}
%%%%%%%%% ABSTRACT
\begin{abstract}
    Images captured in low-light conditions usually suffer from very low contrast, which increases the difficulty of subsequent computer vision tasks in a great extent. In this paper, a low-light image enhancement model based on convolutional neural network and Retinex theory is proposed. Firstly, we show that multi-scale Retinex is equivalent to a feedforward convolutional neural network with different Gaussian convolution kernels. Motivated by this fact, we consider a Convolutional Neural Network(MSR-net) that directly learns an end-to-end mapping between dark and bright images. Different fundamentally from existing approaches, low-light image enhancement in this paper is regarded as a machine learning problem. In this model, most of the parameters are optimized by back-propagation, while the parameters of traditional models depend on the artificial setting. Experiments on a number of challenging images reveal the advantages of our method in comparison with other state-of-the-art methods from the qualitative and quantitative perspective.
\end{abstract}

%%%%%%%%% BODY TEXT
\section{Introduction}

\begin{figure}[t]
  \centering
\subfigure[Origional] {\includegraphics[width=0.32\linewidth]{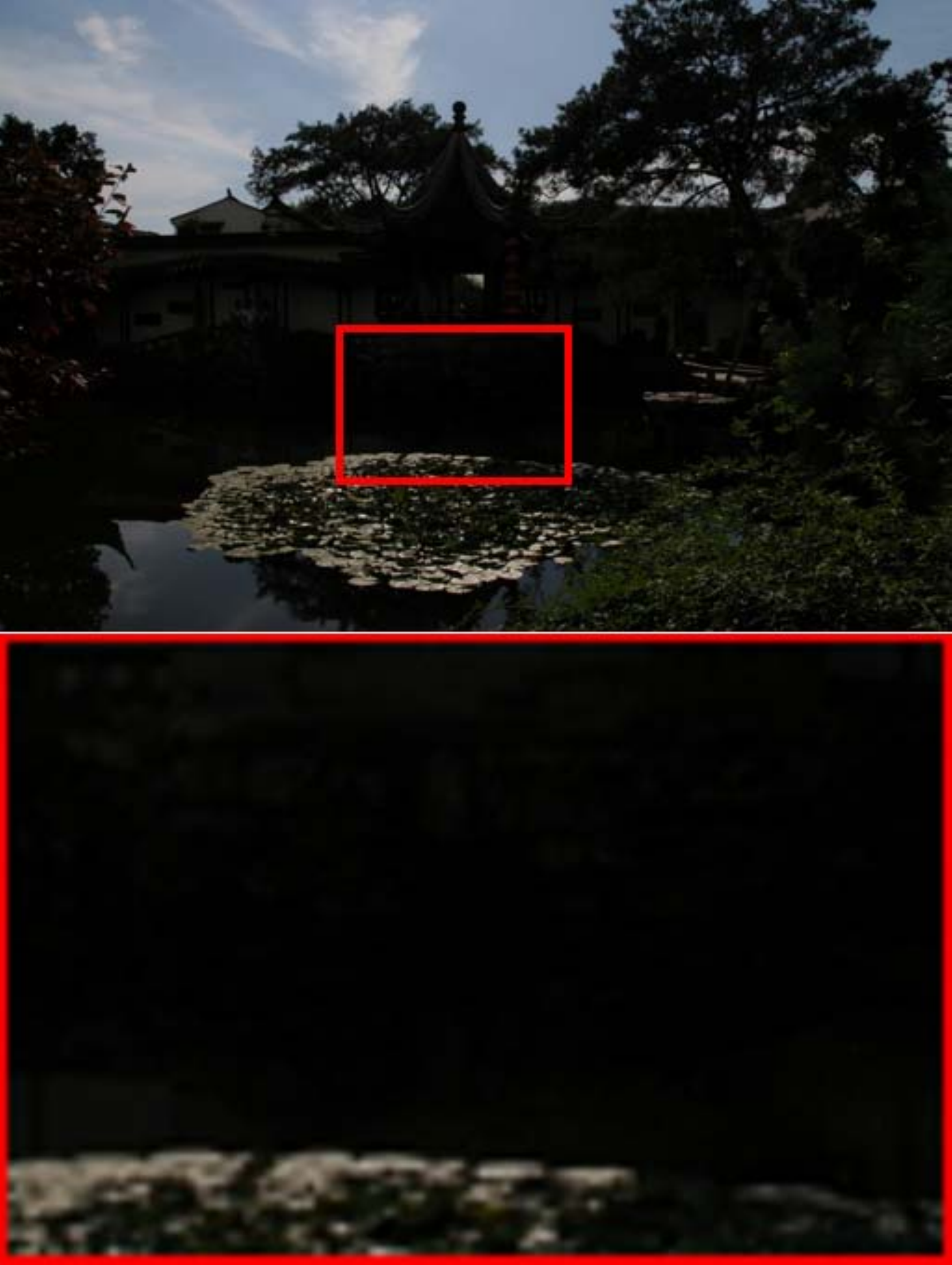}}
\subfigure[MSRCR\cite{jobson1997multiscale}] {\includegraphics[width=0.32\linewidth]{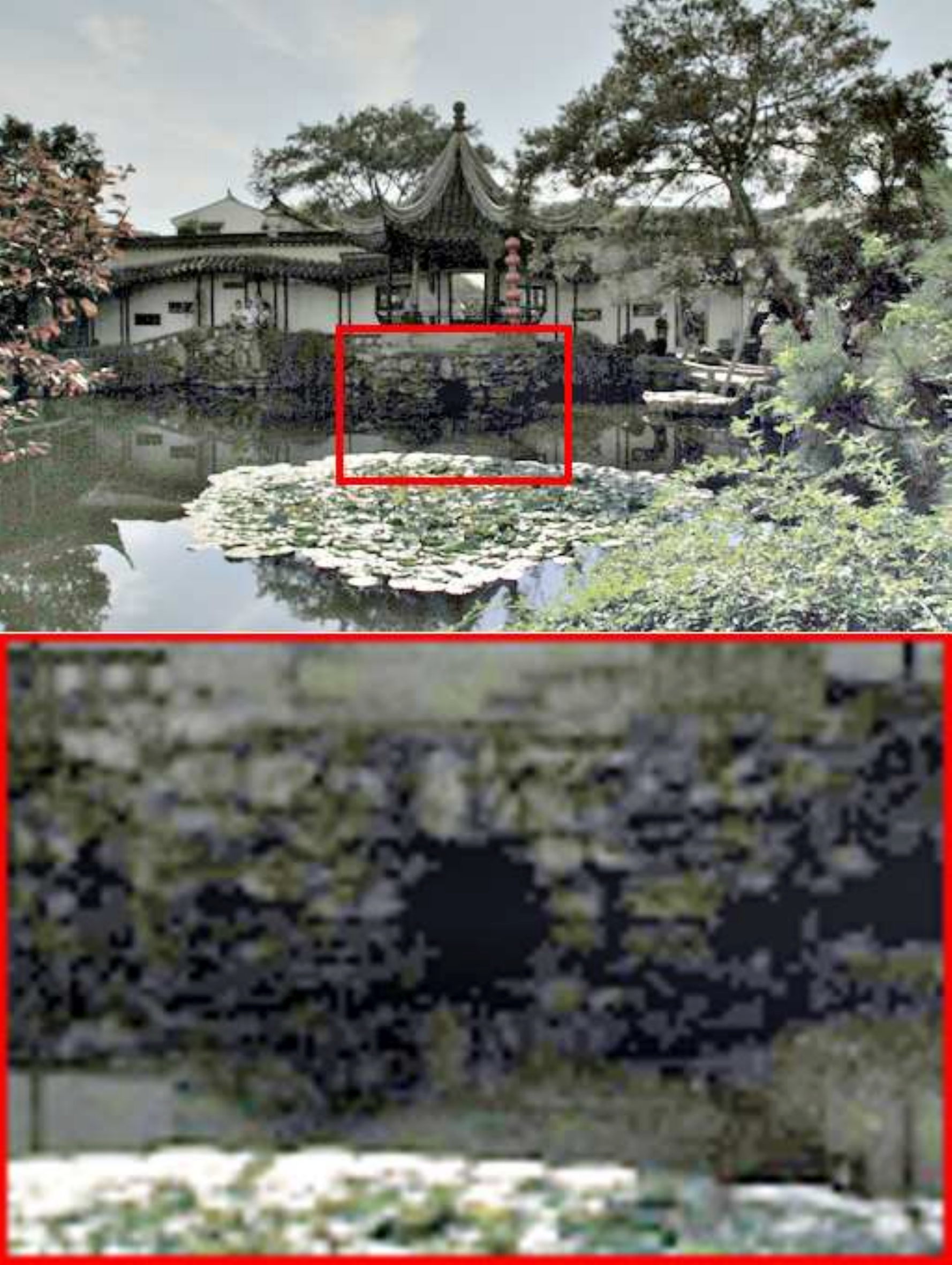}}
\subfigure[Dong\cite{dong2010fast}] {\includegraphics[width=0.32\linewidth]{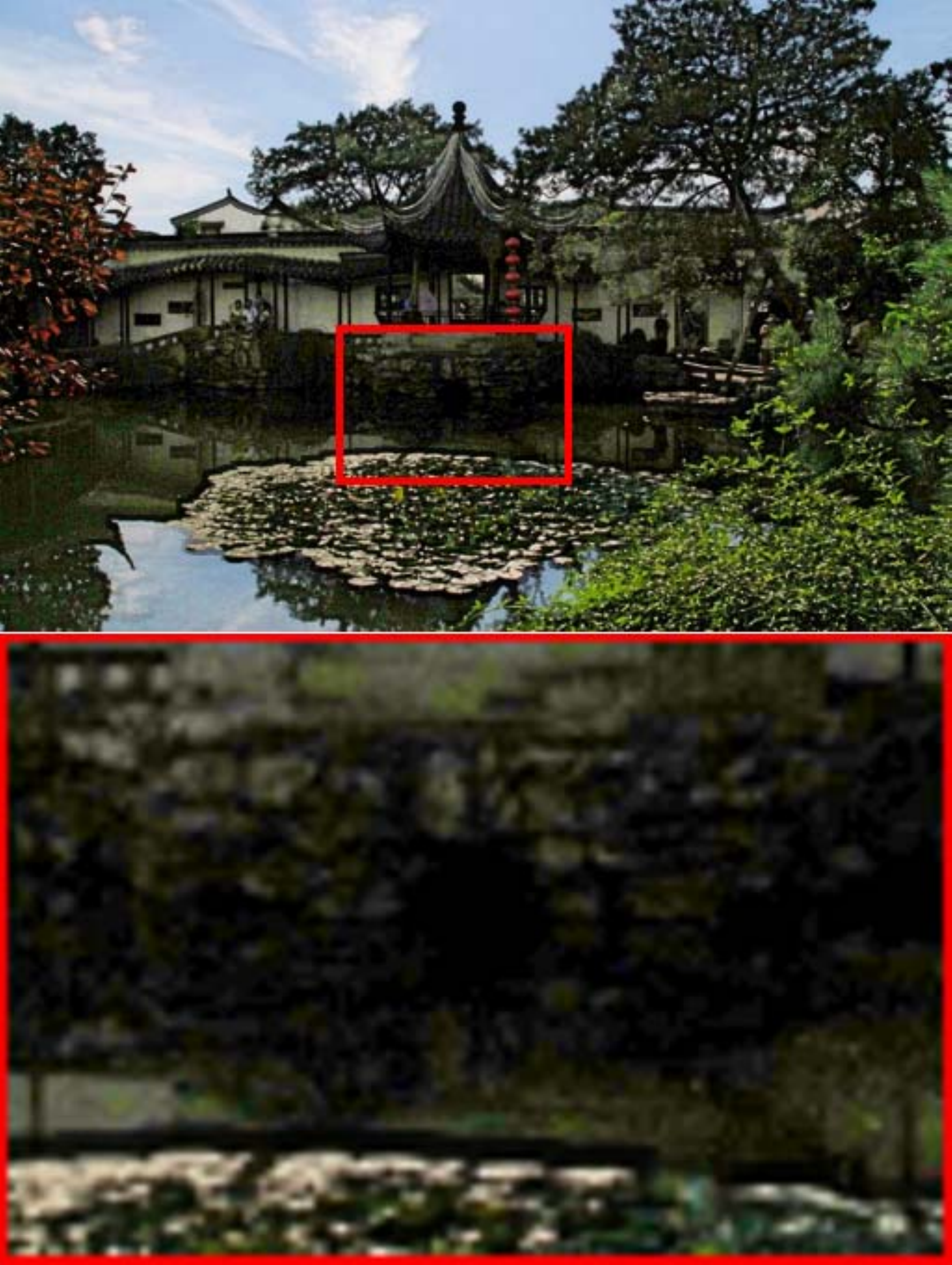}}
\subfigure[LIME\cite{guo2017lime}] {\includegraphics[width=0.32\linewidth]{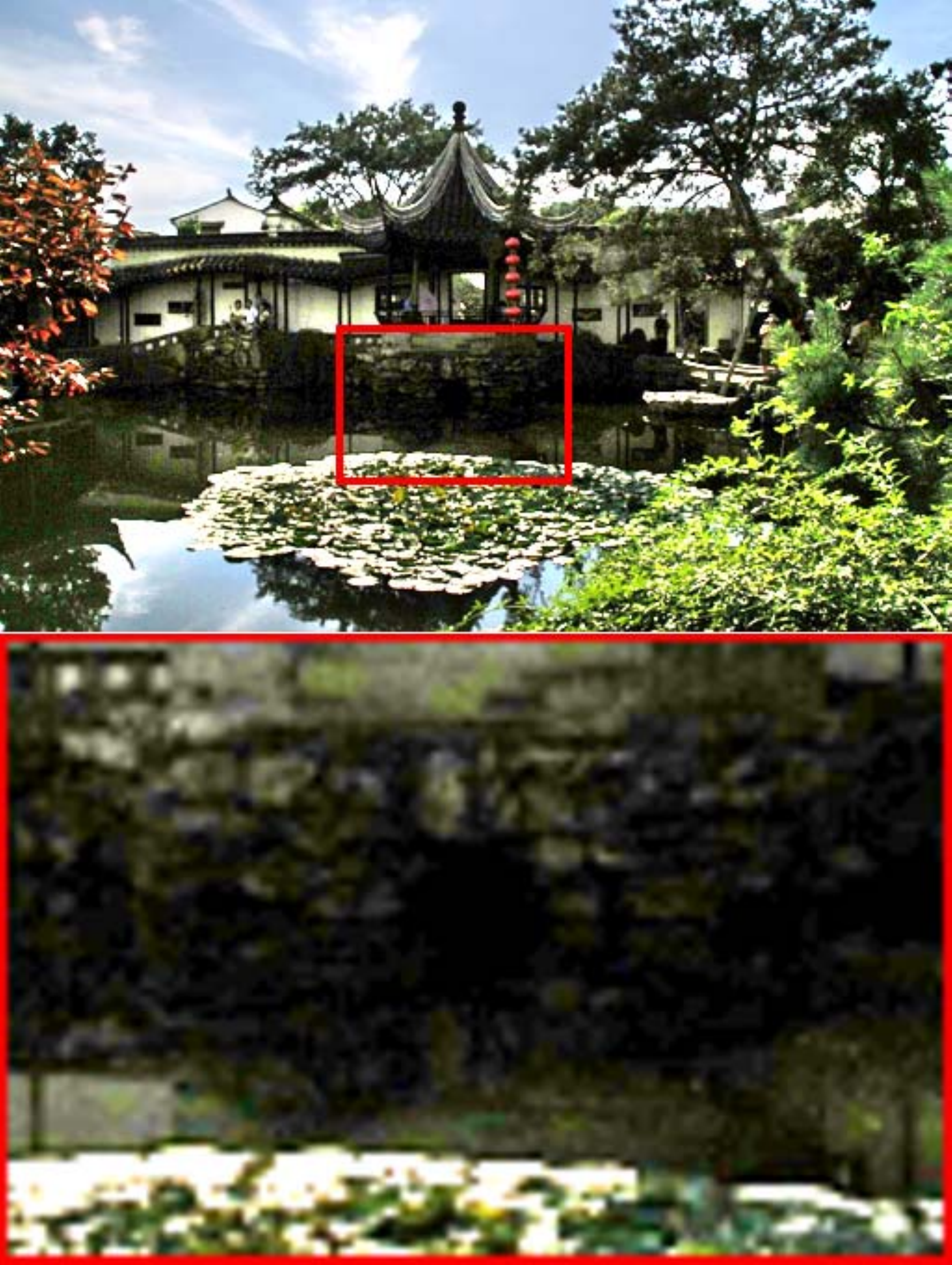}}
\subfigure[SRIE\cite{fu2016weighted}] {\includegraphics[width=0.32\linewidth]{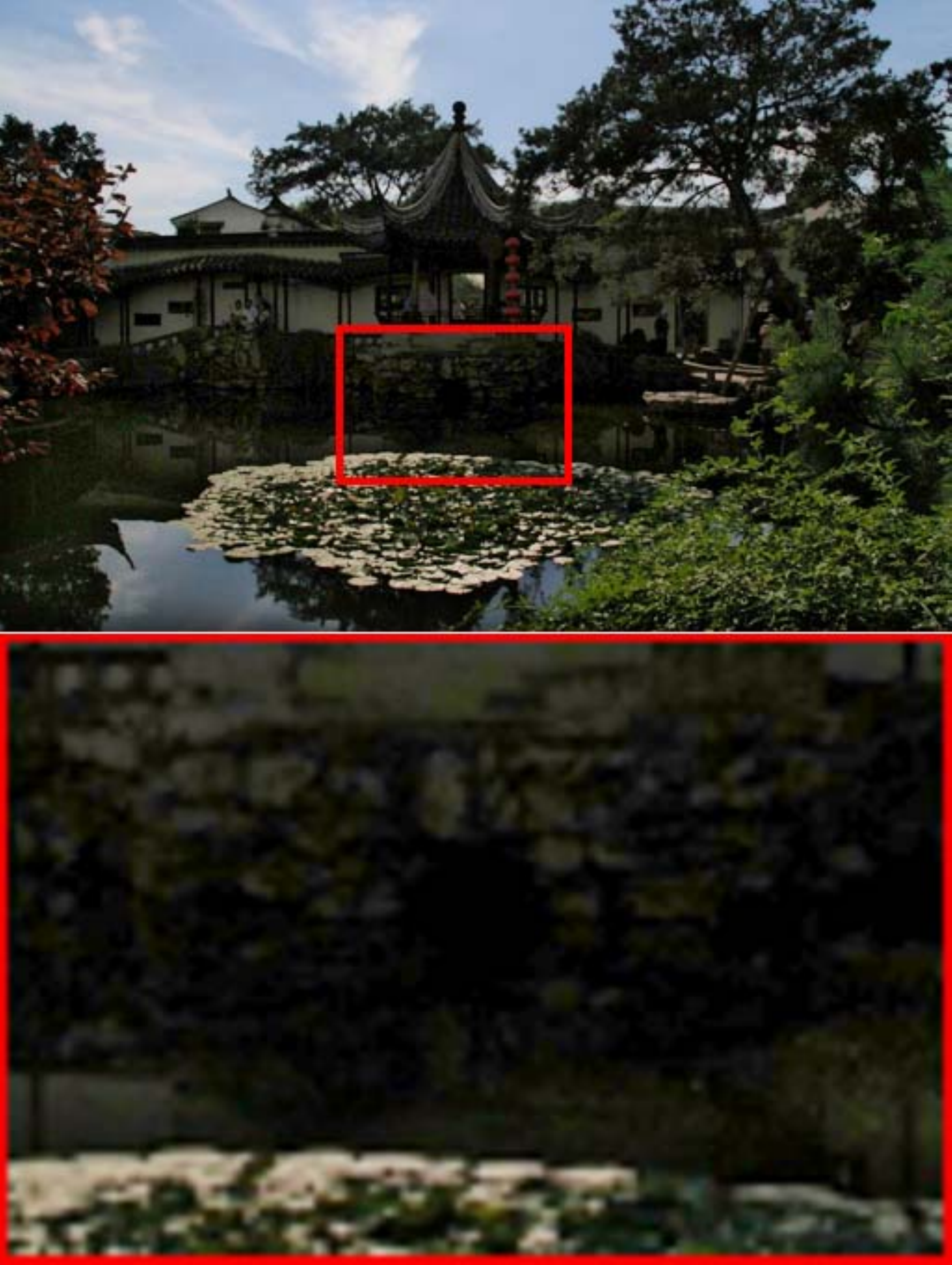}}
\subfigure[Ours] {\includegraphics[width=0.32\linewidth]{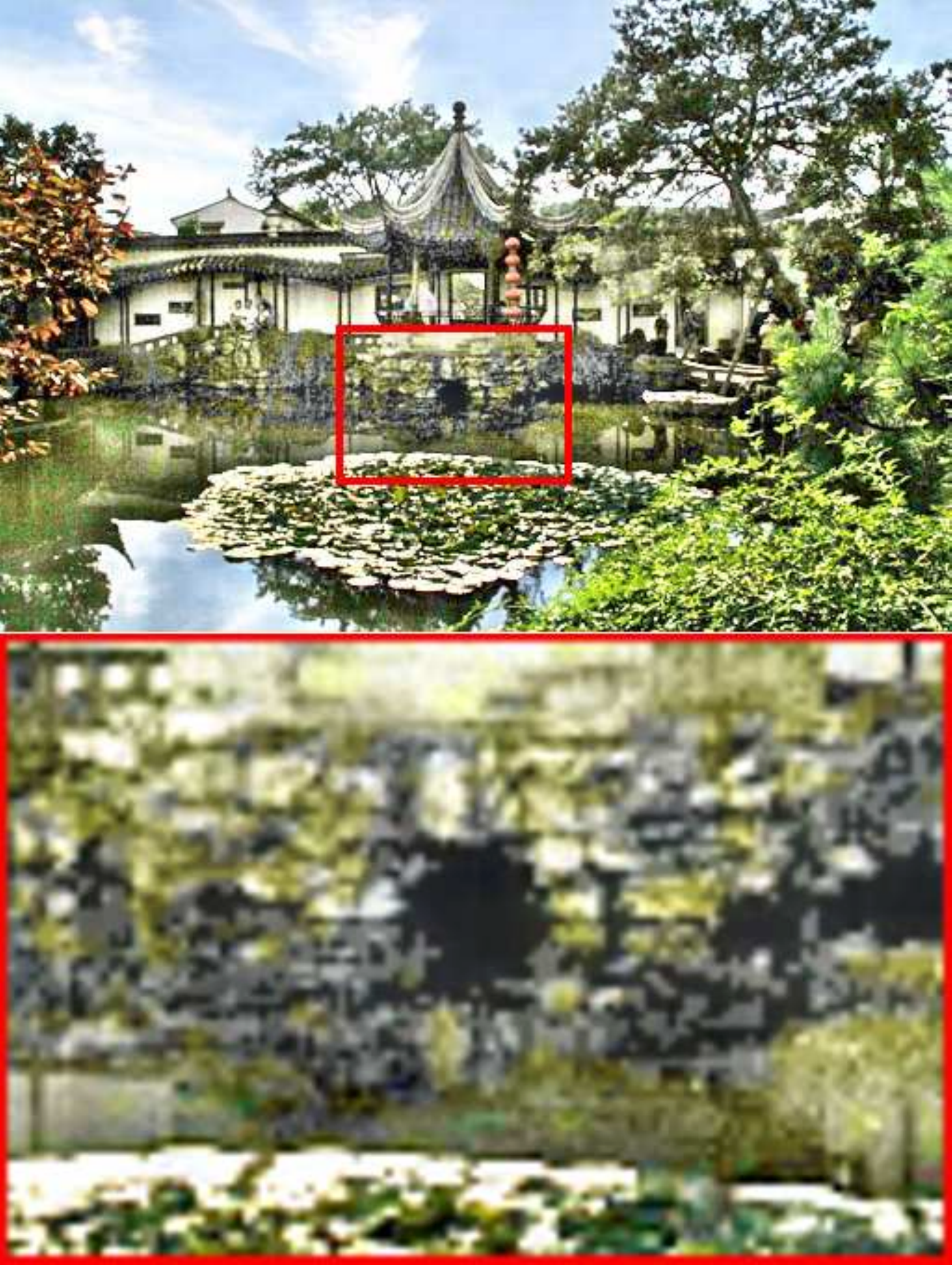}}
  \caption{An example result of our image enhancement method and others state-of-the-art methods.}
  \label{fig:introduction figure}
\end{figure}

There is no doubt that high-quality image plays a critical role in computer vision tasks such as object detection and scene understanding. Unfortunately, the images obtained in reality are often degraded in some cases. For example, when captured in low-light conditions, images always suffer from very low contrast and brightness, which increases the difficulty of subsequent high-level tasks in a great extent. Figure~\ref{fig:introduction figure}(a) provides one case, from which many details have been buried into the dark background. Due to the fact that in many cases only low-light images can be captured, several low-light image enhancement methods have been proposed to overcome this problem. In general, these methods can be categorized into two groups: histogram-based methods and Retinex-based methods.

In this paper, a novel low-light image enhancement model based on convolutional neural network and Retinex theory is proposed. To the best of our knowledge, this is the first work of using convolutional neural network and Retinex theory to solve low-light image enhancement. Firstly, we explain that multi-scale Retinex is equivalent to a feedforward convolutional neural network with different Gaussian convolution kernels. The main drawback of multi-scale Retinex is that the parameters of kernels depend on artificial settings rather than learning from data, which makes the accuracy and flexibility of the model reduce in some way. Motivated by this fact, we put forward a Convolutional Neural Network (MSR-net) that directly learns an end-to-end mapping between dark and bright images. Our method differs fundamentally from existing approaches. We regard low-light image enhancement as a supervised learning problem. Furthermore, the surround functions in Retinex theory~\cite{land1977retinex} are formulated as convolutional layers, which are involved in optimization by back-propagation.

Overall, the contribution of our work can be boiled down to three aspects: First of all, we establish a relationship between multi-scale Retinex and feedforward convolutional neural network. Secondly, we consider low-light image enhancement as a supervised learning problem where dark and bright images are treated as input and output respectively. Last but not least, experiments on a number of challenging images reveal the advantages of our method in comparison with other state-of-the-art methods. Figure~\ref{fig:introduction figure} gives an example. Our method achieves a brighter and more natural result with a clearer texture and richer details.

%-------------------------------------------------------------------------
\section{Related Work}

\subsection{Low-light Image Enhancement}
In general, low-light image enhancement can be categorized into two groups: histogram-based methods and Retinex-based methods.

Directly amplifying the low-light image by histogram transformation is probably the most intuitive way to lighten the dark image. One of the simplest and most widely used technique is histogram equalization(HE), which makes the histogram of the whole image as balanced as possible. Gamma Correction is also a great method to enhance the contrast and brightness by expanding the dark regions and compressing the bright ones in the mean time. However, the main drawback of these method is that each pixel in the image is treated individually, without the dependence of their neighborhoods, which makes the result look inconsistent with real scenes. To resolve the mentioned problems above, variational methods which use different regularization terms on the histogram have been proposed. For example, contextual and variational contrast enhancement~\cite{celik2011contextual} tries to find a histogram mapping to get large gray-level difference.

In this work, Retinex-based methods have been taken into more account. Retinex theory is introduced by Land~\cite{land1977retinex} to explain the color perception property of the human vision system. The dominant assumption of Retinex theory is that the image can be decomposed into reflection and illumination. Single-scale Retinex(SSR)~\cite{jobson1997properties}, based on the center/surround Retinex, is similar to the difference-of-Gaussian(DOG) function which is widely used in natural vision science, and it treats the reflectance as the final enhanced result. Multi-scale Retinex(MSR)~\cite{jobson1997multiscale} can be considered as a weighted sum of several different SSR outputs. However, these methods often look unnatural. Further, modified MSR~\cite{jobson1997multiscale} applies the color restoration function(CRF) in the chromaticity space to eliminate the color distortions and gray zones evident in the MSR output. Recently, the method proposed in~\cite{guo2017lime} tries to estimate the illumination of each pixel by finding the maximum value in R, G and B channel, then refines the initial illumination map by imposing a structure prior on it. Seonhee Park~\etal\cite{park2017low} use the variational-optimization-based Retinex algorithm to enhance the low-light image. Fu~\etal\cite{fu2016weighted} propose a new weighted variational model to estimate both the reflection and the illumination. Different from conventional variational models, their model can preserve the estimated reflectance with more details. Inspired by the dark channel method on de-hazing,~\cite{dong2010fast} finds the inverted low-light image looks like haze image. They try to remove the inverted low-light image of haze by using the method proposed in~\cite{he2011single} and then invert it again to get the final result.
\subsection{Convolutional Neural Network for Low-level Vision Tasks}

Recently, powerful capability of deep neural network~\cite{krizhevsky2012imagenet} has led to dramatic improvements in object recognition~\cite{he2016deep}, object detection~\cite{ren2015faster}, object tracking~\cite{wang2015visual}, semantic segmentation~\cite{long2015fully} and so on. Besides these high-level vision tasks, deep learning has also shown great ability at low-level vision tasks. For instance, Dong~\etal\cite{dong2016image} train a deep convolutional neural network (SRCNN) to accomplish the image super-resolution tasks. Fu~\etal\cite{furemoving} try to remove rain from single images via a deep detail network. Cai~\etal\cite{cai2016dehazenet} propose a trainable end-to-end system named DehazeNet, which takes a hazy image as input and outputs its medium transmission map that is subsequently used to recover a haze-free image via atmospheric scattering model.
%-------------------------------------------------------------------------
\section{CNN Network for Low-light Image Enhancement}

\begin{figure*}[!t]
\centering
\subfigure[MSR architecture] {\includegraphics[width=0.7\linewidth]{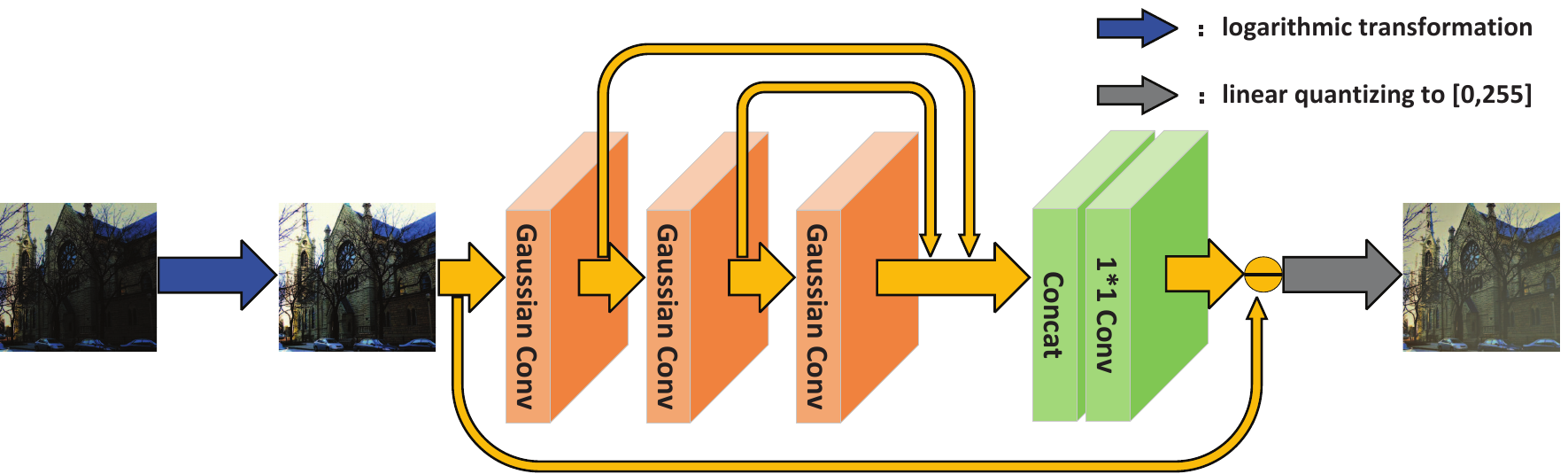}}
\subfigure[MSR-net architecture] {\includegraphics[width=0.85\linewidth]{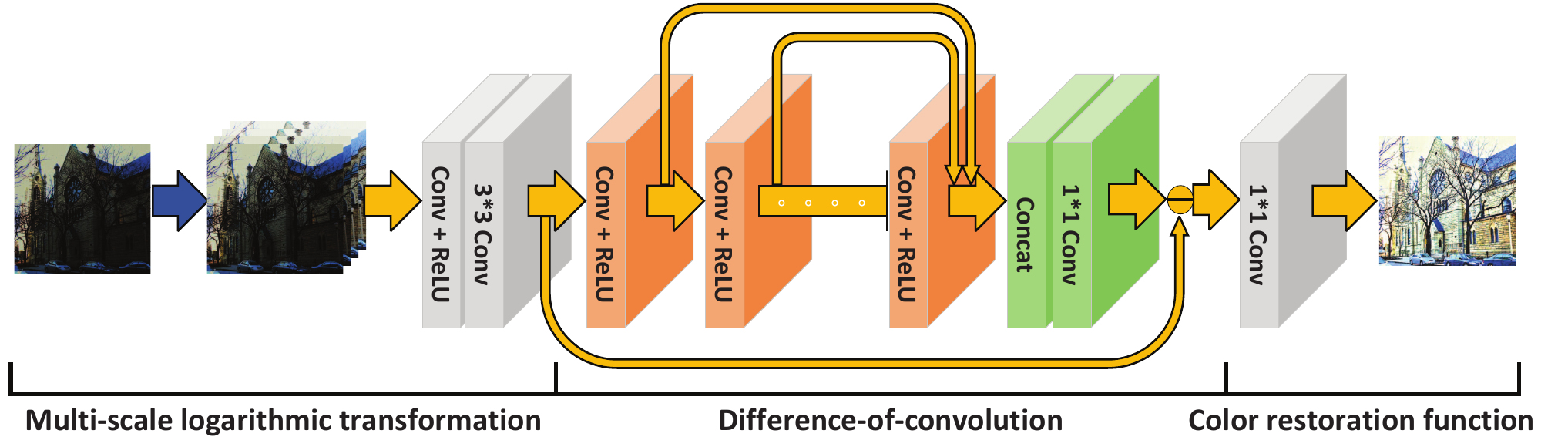}}
\caption{The architecture of MSR and our proposed MSR-net. Obviously, MSR-net is similiar to the MSR to some extent. In such a network, it is divided into three parts, Multi-scale Logarithmic Transformation, Difference-of-convolution and Color Restoration Function.}
\label{fig:MSR and MSR-net flow chart}
\end{figure*}

We elaborate that multi-scale Retinex as a low-light image enhancement method is equivalent to a feedforward convolutional neural network with different Gaussian convolution kernels from a novel perspective. Subsequently, we propose a Convolutional Neural Network (MSR-net) that directly learns an end-to-end mapping between dark and bright images.

\subsection{Multi-scale Retinex is a CNN Network}\label{Multi-scale Retinex is a CNN network}
The dominant assumption of Retinex theory is that the image can be decomposed into reflection and illumination:
\begin{equation}
{I}(x,y) = {r}(x,y) \cdot {S}(x,y)
\end{equation}

Where $I$ and $r$ represent the captured image and the desired recovery, respectively. Single-scale Retinex(SSR)~\cite{jobson1997properties}, based on the center/surround Retinex, is similar to the difference-of-Gaussian(DOG) function which is widely used in natural vision science. Mathematically, this takes the form
\begin{equation}
{R_i}(x,y) = \log {I_i}(x,y) - \log \left[ {F(x,y) * {I_i}(x,y)} \right]\label{equ:log1}
\end{equation}

Where ${R_i}(x,y)$ is the associated Retinex output, ${I_i}(x,y)$ is the image distribution in the ${i^{th}}$ color spectral band, ${*}$ denotes the convolution operation, and $F(x,y)$ is the Gaussian surround function
\begin{equation}
F(x,y) = K{e^{ - \frac{{{x^2} + {y^2}}}{{{2c^2}}}}}
\end{equation}

Where ${c}$ is the standard deviation of Gaussian function, and ${K}$ is selected such that
\begin{equation}
\iint {F(x,y)dxdy = 1}
\end{equation}

By changing the position of the logarithm in the above formula and setting
\begin{equation}
{R_i}(x,y) = \log {I_i}(x,y) - \left[ {\log {I_i}(x,y)} \right] * F(x,y)\label{equ:log2}
\end{equation}

In this way we obtained a classic high pass linear filter, but applied to $\log I$ instead of ${I}$. The above two formulas, Equation~\ref{equ:log1} and Equation~\ref{equ:log2}, are of course not equivalent in mathematical form. The former is the logarithm of ratio between the image and a weighted average of it, while the latter is the logarithm of ratio between the image and a weighted product. Actually, this amounts to choosing between an arithmetic mean and a geometric mean. Experiments show that these two methods are not much different. In this work we choose the latter for simplicity.

Further, multi-scale Retinex(MSR)~\cite{jobson1997multiscale} is considered as a weighted sum of the outputs of several different SSR outputs. Mathematically,
\begin{equation}
{R_{MSR_{i}}} = \sum\limits_{n = 1}^N {{w_n}{R_{n_i}}}
\end{equation}

Where ${N}$ is the number of scales, ${R_{n_i}}$ denotes the ${i^{th}}$ component of the ${n^{th}}$ scale, ${R_{MSR_{i}}}$ represents the ${i^{th}}$ spectral component of the MSR output and ${w_n}$ is the weight associated with the ${n^{th}}$ scale.

After experimenting with one small scale (standard deviation $c < 20$) and one large scale (standard deviation $c >200$), the need for the third intermediate scale is immediately apparent in order to eliminate the visible ``halo" artifacts near strong edges~\cite{jobson1997multiscale}. Thus, the formula is as follows:
\begin{equation}
\begin{aligned}
&{R_{MSR_{i}}}(x,y)=\\&\frac{1}{3}\sum\limits_{n = 1}^3 {\left\{ {\log {I_i}(x,y) - \left[ {\log {I_i}(x,y)} \right] * {F_n}(x,y)} \right\}}
\end{aligned}
\end{equation}

More concrete, we have
\begin{equation}
\begin{aligned}
&{R_{MSR_{i}}}(x,y) = \\&\log {I_i}(x,y) - \frac{1}{3}\log {I_i}(x,y) * \left[ {\sum\limits_{n = 1}^3 {{K_n}{e^{ - \frac{{{x^2} + {y^2}}}{{2c_n^2}}}}} } \right]\label{equ:Gaussian}
\end{aligned}
\end{equation}

Noticing the fact that convolution of two Gaussian functions is still a Gaussian function, whose variance is equal to the sum of two original variance. Therefore, we can represent the above equation~\ref{equ:Gaussian} by using the cascading structure, as Figure~\ref{fig:MSR and MSR-net flow chart}(a) shows.

The three cascading convolution layers are considered as three different Gaussian kernels. More concrete, the parameter of the first convolution layer is based on a Gaussian distribution, whose variance is $c_1^2$. Similarly, the variances of the second and the third convolution layers are $c_2^2 - c_1^2$, $c_3^2 - c_2^2$, respectively. At last, the concatenation and $1 \times 1$ convolution layers represent the weighted average. In a word, multi-scale Retinex is practically equivalent to a feedforward convolutional neural network with a residual structure.

\subsection{Proposed Method}\label{section:Proposed method}

In the previous section, we put forward the fact that multi-scale Retinex is equivalent to a feedforward convolutional neural network. In this section, inspired by the novel fact, we consider a convolutional neural network to solve the low-light image enhancement problem. Our method outlined in Figure~\ref{fig:MSR and MSR-net flow chart}(b) differs fundamentally from existing approaches, which takes low-light image enhancement as a supervised learning problem. The input and output data correspond to the low-light and bright images, respectively. More detail about our training dataset will be explained in section~\ref{section:experiments}.

Our model consists of three components: \textbf{Multi-scale Logarithmic Transformation}, \textbf{Difference-of-convolution} and \textbf{Color Restoration Function}. Compared to single-scale logarithmic transformation in MSR, our model attempts to use multi-scale logarithmic transformation, which has been verified to achieve a better performance in practice. Figure~\ref{fig:Comparison with N} gives an example. Difference-of-convolution plays an analogous role with difference-of-Gaussian in MSR, and so does color restoration function. The main difference between our model and original MSR is that most of the parameters in our model are learned from the training data, while the parameters in MSR such as the variance and other constant depend on the artificial setting.

Formally, we denote the low-light image as input ${X}$ and corresponding bright image as ${Y}$. Suppose ${f_1}$, ${f_2}$, ${f_3}$ denote three sub-functions: multi-scale logarithmic transformation, difference-of-convolution, and color restoration function. Our model can be written as the composition of three functions:
\begin{equation}
  f(X) = {f_3}({f_2}({f_1}(X)))
\end{equation}

\textbf{Multi-scale Logarithmic Transformation:} Multi-scale logarithmic transformation ${f_1}(X)$ takes the original low-light image $X$ as input and computes the same size output ${X_1}$. Firstly, the dark image is enhanced by several difference logarithmic transformation. The formula is as follows:
\begin{equation}
  {M_j} = {\log _{{v_j} + 1}}(1 + {v_j} \cdot X),j = 1,2,...,n
\end{equation}

Where $M_j$ denotes the output of the $j^{th}$ scale with the logarithmic base $v_j+1$ , and $n$ denotes the number of logarithmic transformation function. Next, we concatenate these 3D tensors $M_j$ (3 channels$\times$ width $\times$ height) to a larger 3D tensor $M$ (3n channels $\times$ width $\times$ height) and then make it go through convolutional and ReLU layers.
\begin{flalign}
&&M = &\left[ {{M_1},{M_2},...,{M_n}} \right] &\\
&&{X_1} =& \max (0,M * {W_{ - 1}} + {b_{ - 1}}) * {W_0} + {b_0}
\end{flalign}

Where * denotes a convolution operator, $W_{-1}$ is a convolution kernel that shrinks the $3n$ channels to 3 channels, $\max (0, \cdot )$ corresponds to a ReLU and $W_0$ is a convolution kernel with three output channels for better nonlinear representation. As we can see from the above operation, this part is mainly designed to get a better image via weighted sums of multiple logarithmic transformations, which accelerates the convergence of the network.

\textbf{Difference-of-convolution:} Difference-of-convolution function $f_2$ takes the input $X_1$ and computes the same size output $X_2$. Firstly, the input $X_1$ passes through multi-convolutional layers.
\begin{flalign}
&&H_0 = & X_1 &\\
&&{H_m} = & \max (0,{H_{m - 1}} * {W_m} + {b_m}),m = 1,2,...,K
\end{flalign}

Where $m$ denotes the $m^{th}$ convolutional layer, $K$ is equal to the number of convolutional layers. And $W_m$ represents the  $m^{th}$ kernel. As mentioned earlier in section~\ref{Multi-scale Retinex is a CNN network}, $H_1,H_2,...,H_K$ are considered as smooth images at different scales, then we concatenate these 3D tensors $H_m$ to a larger 3D tensor $H$ and get it pass the convolutional layer:
\begin{flalign}
  &H = [H_1,H_2,...,H_K] \\
  &{H_{K+1}} = H*W_{K+1}+b_{K+1}
\end{flalign}

Where the $W_{K+1}$ is a convolutional layer with three output channels and the $1 \times 1$  receptive field, which is equivalent to averaging these $K$ images. Similar to MSR, the output of $f_2$ is the subtraction between  $X_1$ and $H_{K+1}$:
\begin{equation}
  {X_2} = {f_2}({X_1}) = {X_1} - {H_{K + 1}}
\end{equation}

\textbf{Color Restoration Function:} Considering that MSR result often looks unnatural, modified MSR~\cite{jobson1997multiscale} applies the color restoration function(CRF) in the chromaticity space to eliminate the color distortions and gray zones evident in the MSR output. In our model CRF is imitated by a $1 \times 1$ convolutional layer with three output channels:
\begin{equation}
  \hat Y = {f_3}({X_2}) = {X_2} * {W_{K + 2}} + {b_{K + 2}}
\end{equation}

\begin{figure}[ht]
\centering
\subfigure[Input] {\includegraphics[width=0.45\linewidth,height=3cm]{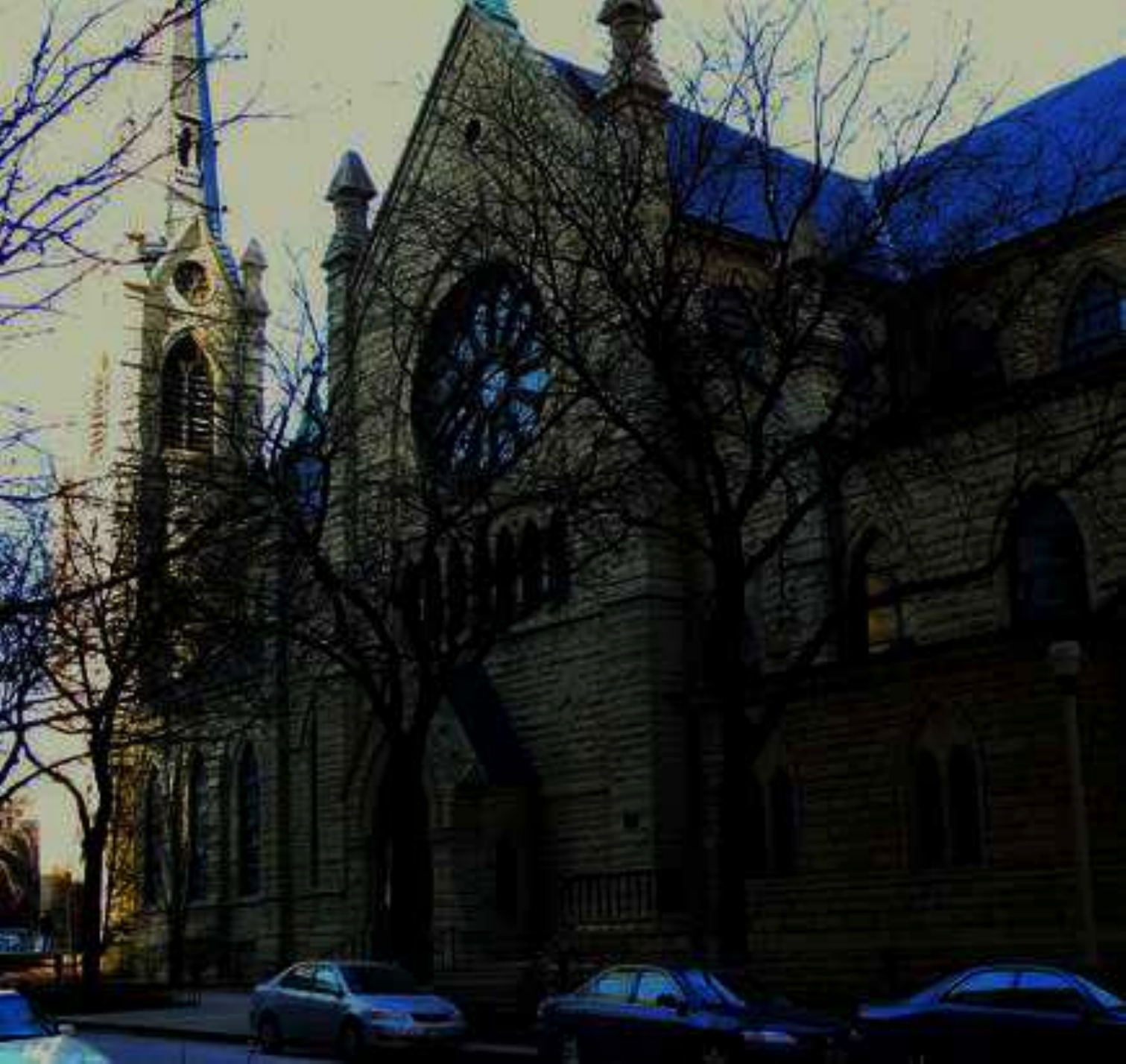}}
\subfigure[The result of $f_1$] {\includegraphics[width=0.45\linewidth,height=3cm]{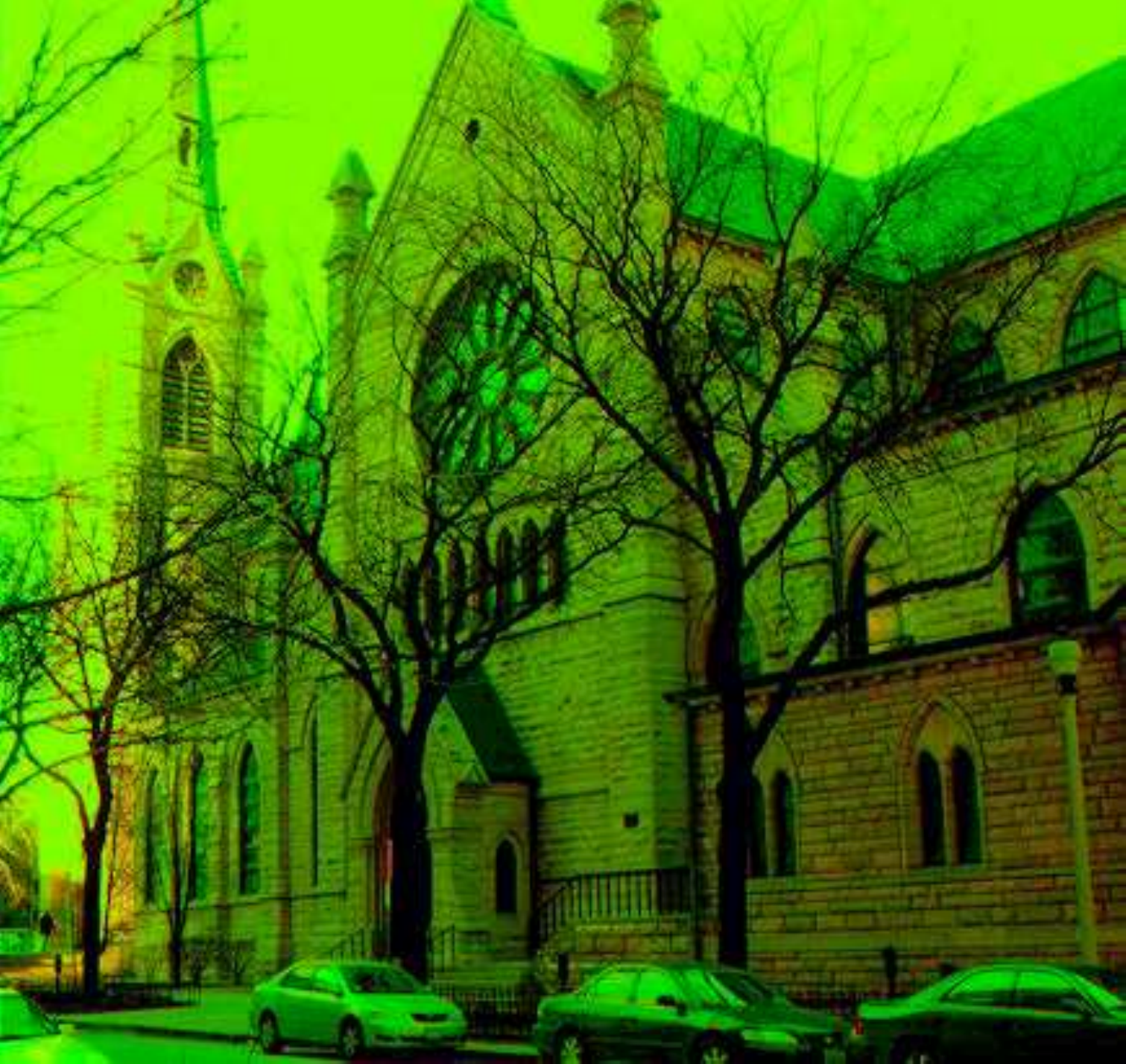}}
\subfigure[The result of $f_2$] {\includegraphics[width=0.45\linewidth,height=3cm]{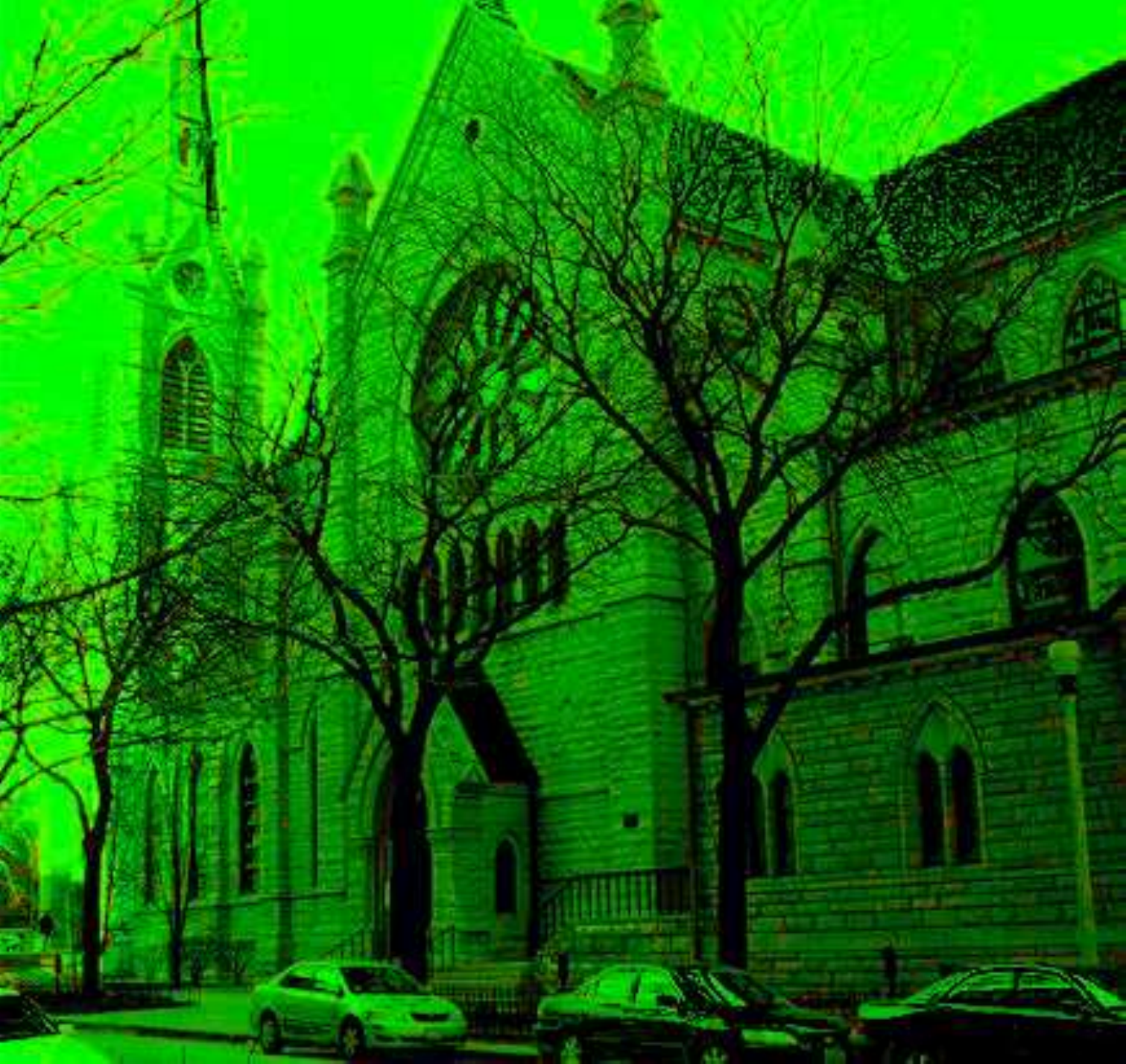}}
\subfigure[The result of $f_3$] {\includegraphics[width=0.45\linewidth,height=3cm]{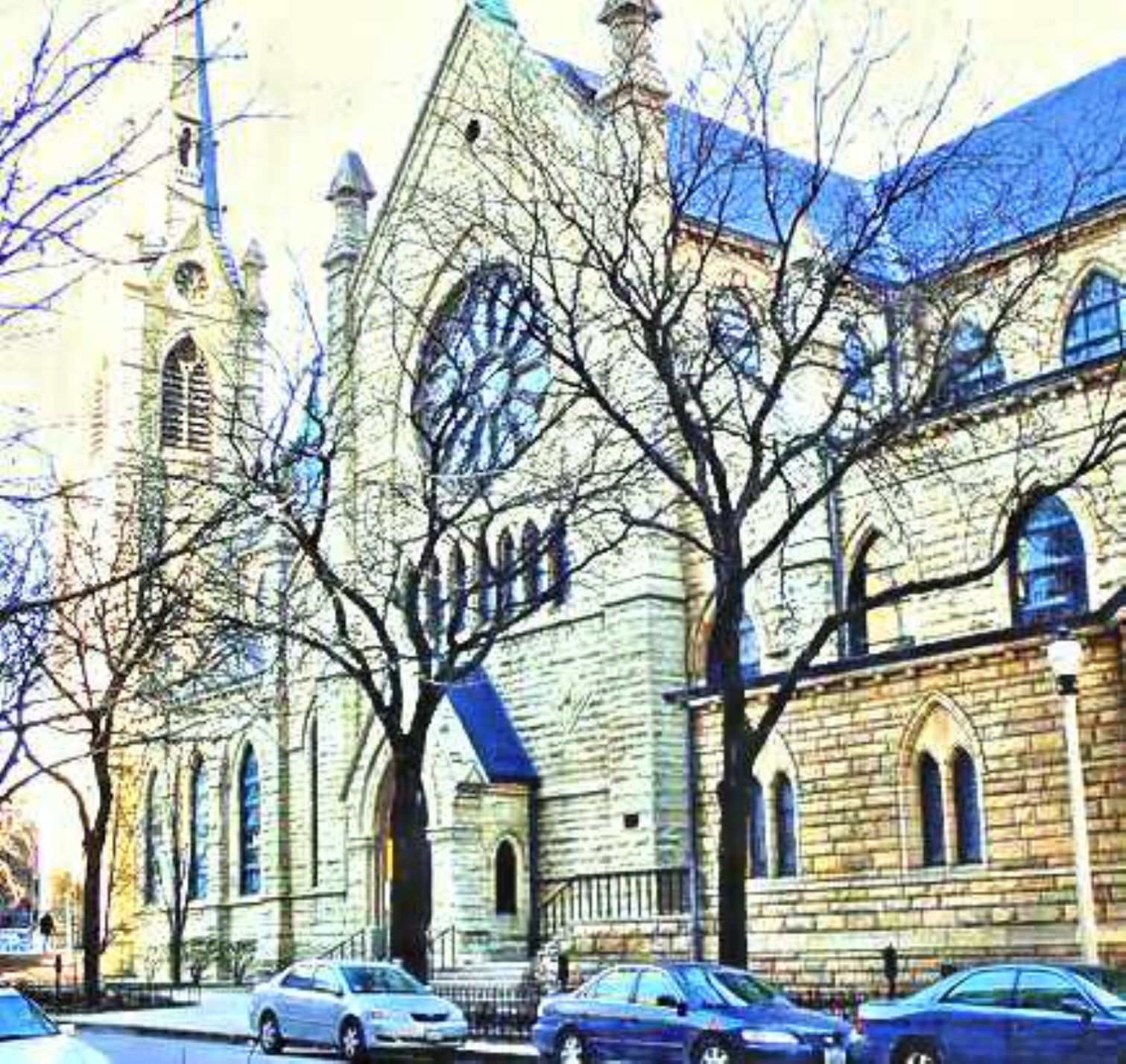}}
\caption{ A low light image and the results of $f_1$,$f_2$,$f_3$}
\label{fig:results of different stages}
\end{figure}

Where $\hat Y$ is the final enhanced image. For more visualization, a low light image and the results of $f_1$,$f_2$,$f_3$ have been shown in Figure~\ref{fig:results of different stages} respectively.

\subsection{Objective function }
The goal of our model is to train a deep convolutional neural network to make the output $f(X)$ and the label ${Y}$ as close as possible under the criteria of Frobenius norm.
\begin{equation}\
  L = \frac{1}{N}\sum\limits_{i = 1}^N {\left\| {f({X_i}) - {Y_i}} \right\|_F^2}+{\lambda}\sum\limits_{i = -1}^{K+2} {\left\| {W_i} \right\|_F^2}
\end{equation}

Where $N$ is the number of training samples, $\lambda$ represents the regularization parameter.

Weights $W$ and bias $b$ are the whole parameters in our model. Besides, the regularization parameter $\lambda$, the number of logarithmic transformation function $n$, the scale of logarithmic transformation $v$ and the number of convolutional layers $K$, are considered as the hyper-parameters in the model. The parameters in our model are optimized by back-propagation, while the hyper-parameters are chosen by grid-search. More detail about the sensitivity analysis of hyper-parameters will be elaborated in section~\ref{section:experiments}.
\begin{figure*}[!t]
\centering
\subfigure[Ground truth] {\includegraphics[width=0.135\linewidth]{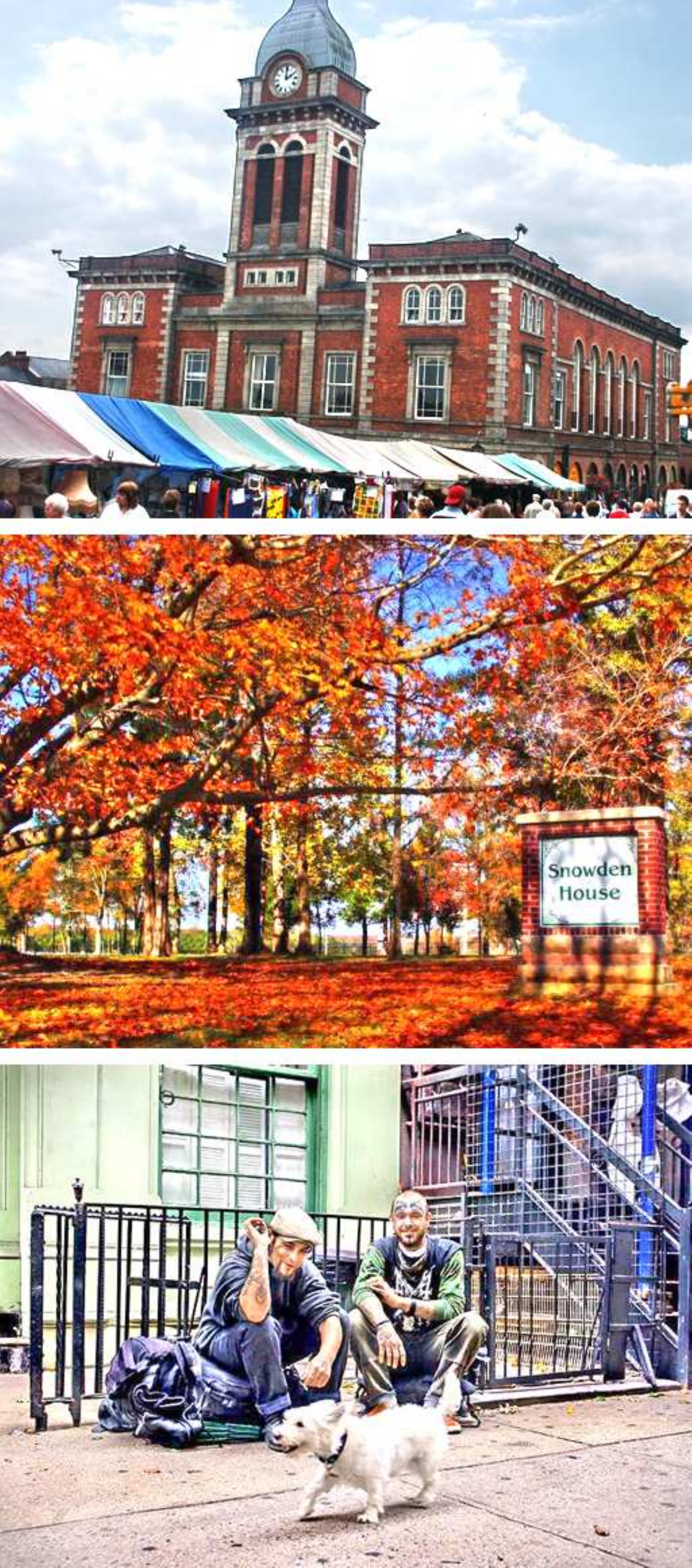}}
\subfigure[Synthetic data] {\includegraphics[width=0.135\linewidth]{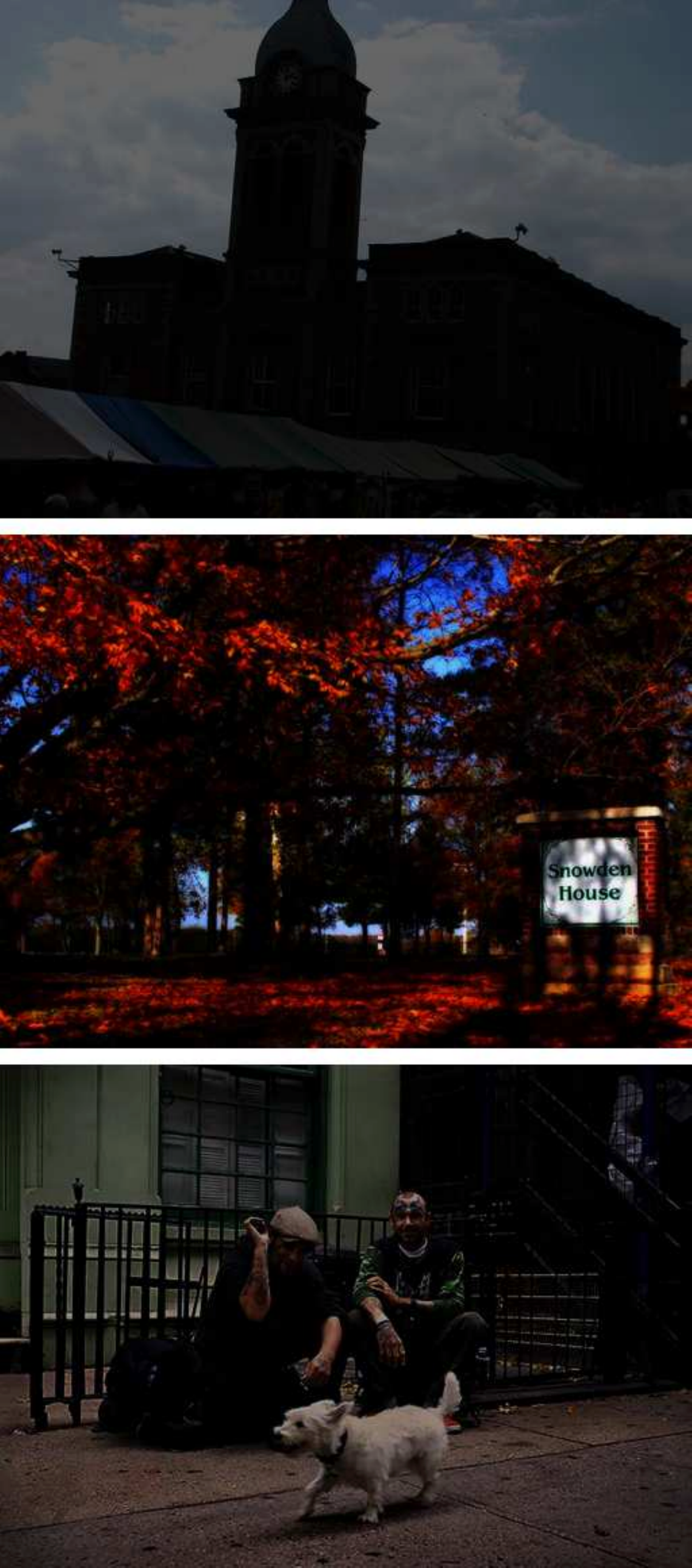}}
\subfigure[MSRCR\cite{jobson1997multiscale}] {\includegraphics[width=0.135\linewidth]{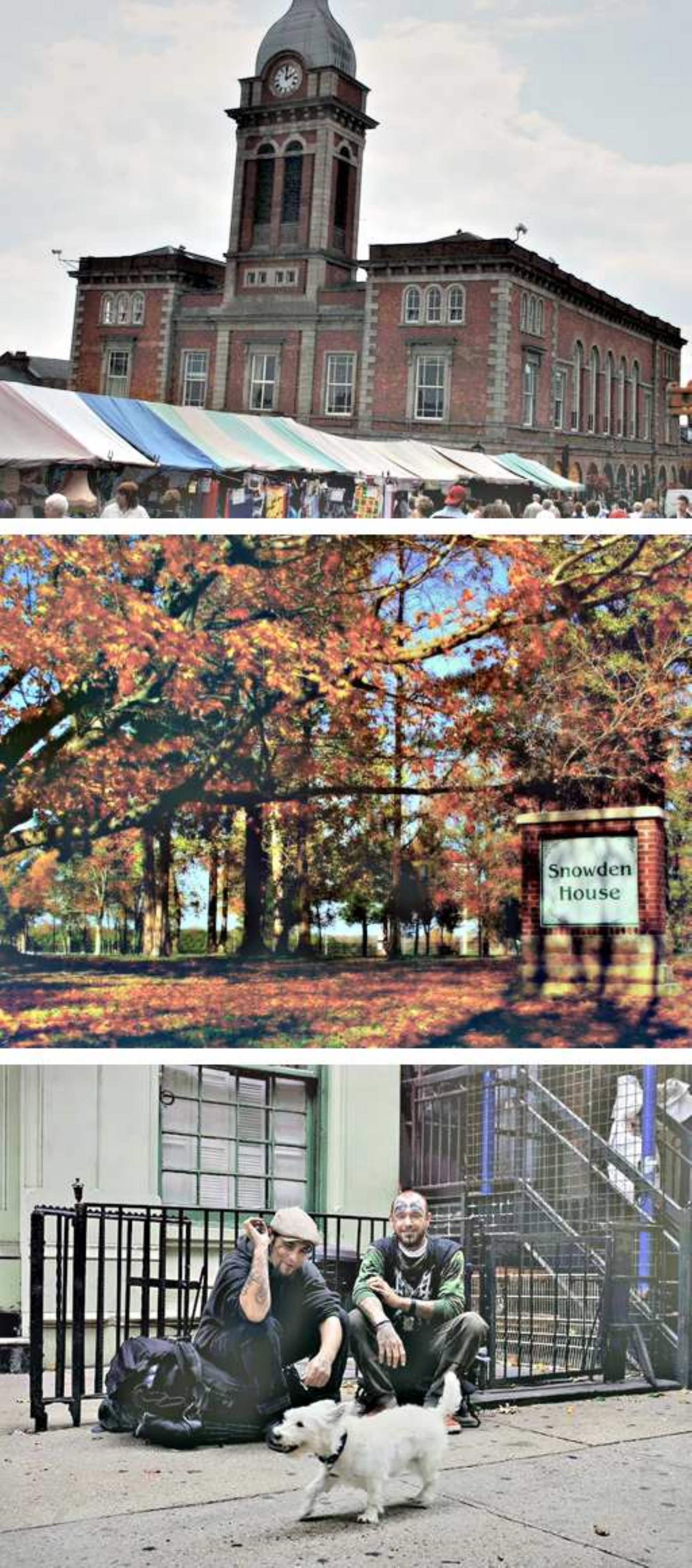}}
\subfigure[Dong\cite{dong2010fast}] {\includegraphics[width=0.135\linewidth]{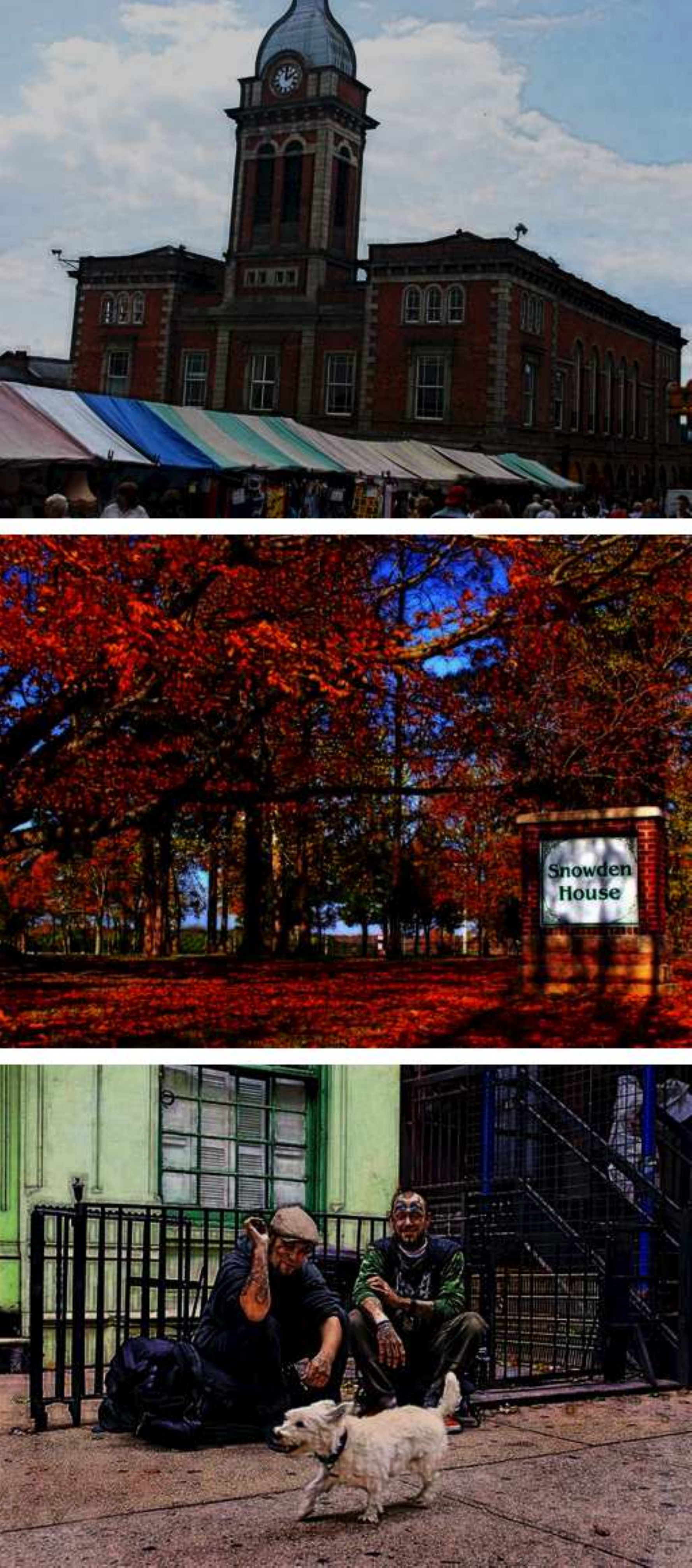}}
\subfigure[LIME\cite{guo2017lime}] {\includegraphics[width=0.135\linewidth]{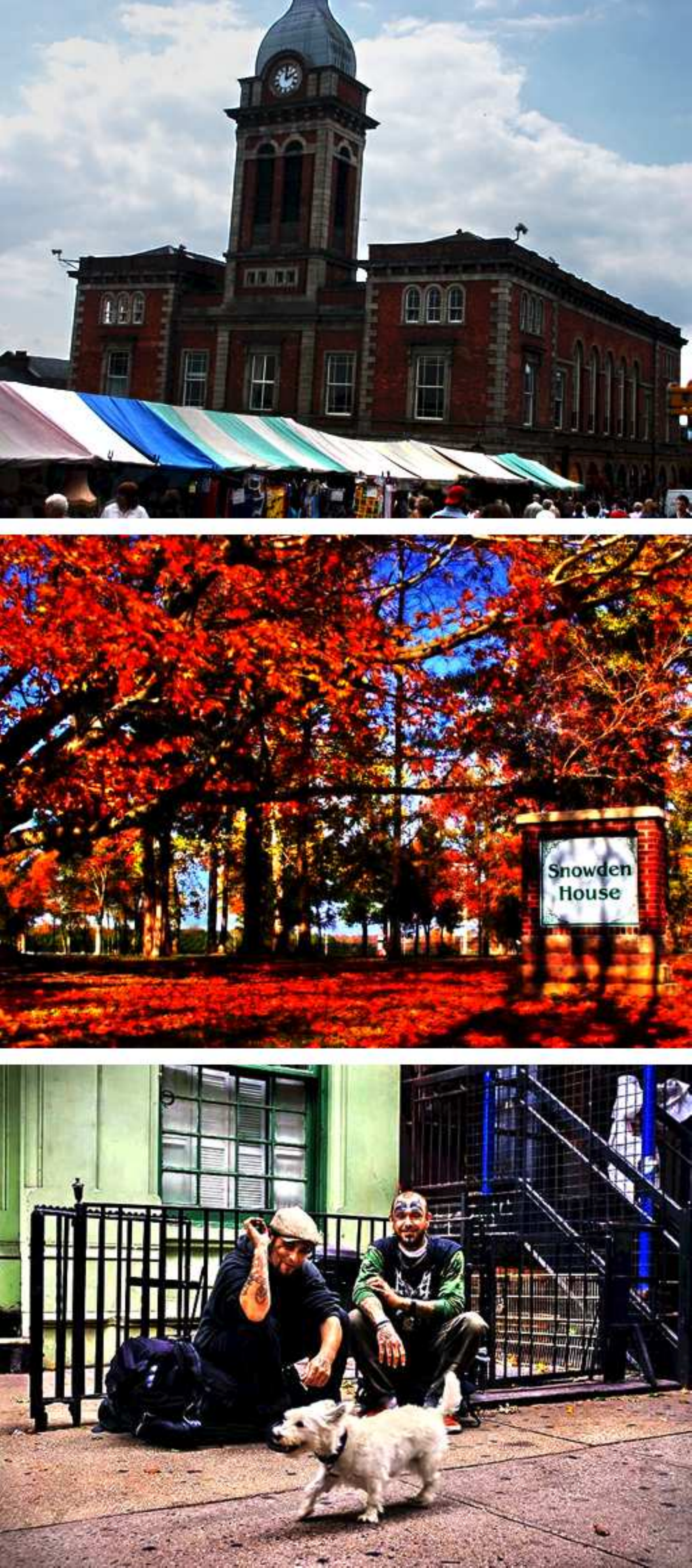}}
\subfigure[SRIE\cite{fu2016weighted}] {\includegraphics[width=0.135\linewidth]{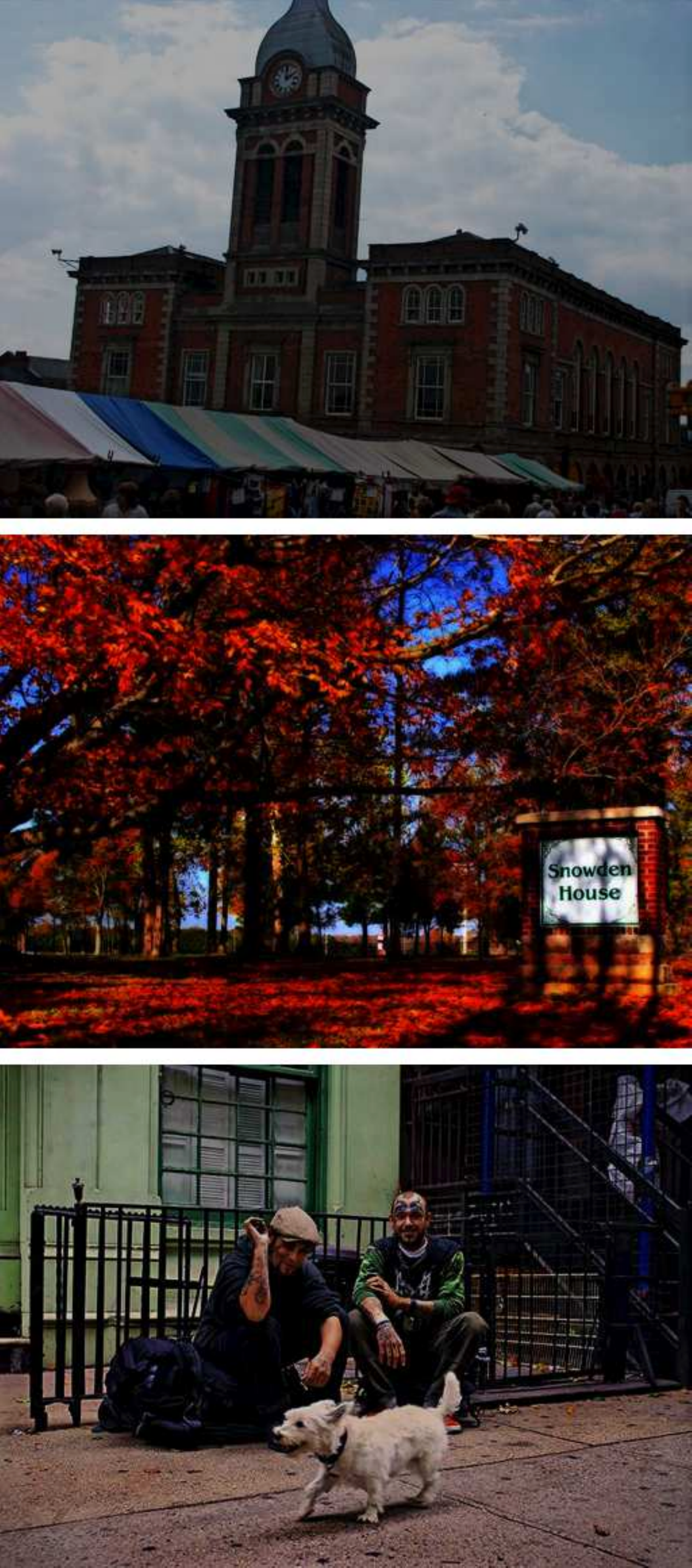}}
\subfigure[Ours] {\includegraphics[width=0.135\linewidth]{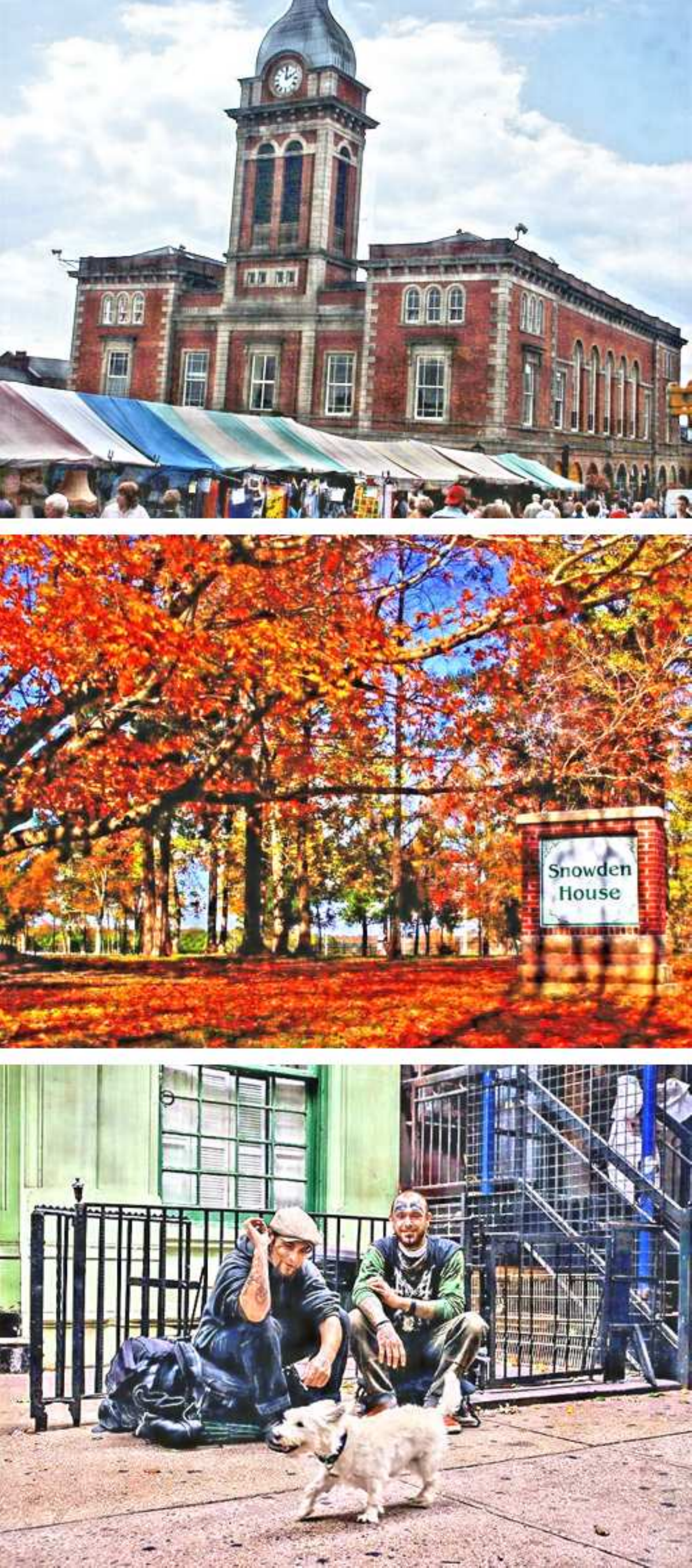}}
\caption{Results using different methods on synthesized test images}
\label{fig:synthesized test image}
\end{figure*}

\begin{table*}[tbp]
\centering  % 表居中
\caption{ Quantitative measurement results using SSIM/NIQE on synthesized test images}\label{tab:synthesized test images}
\begin{tabular}{c|ccccccc}  % {lccc} 表示各列元素对齐方式，left-l,right-r,center-c
\hline
Dataset     &Ground truth   &Synthetic image    & MSRCR\cite{jobson1997multiscale}   &Dong\cite{dong2010fast}     & LIME\cite{guo2017lime}    &SRIE\cite{fu2016weighted}    & Ours \\ \hline  % \hline 在此
1st row     &1/2.69         &0.35/6.44          &0.89/3.29  &0.64/3.57  &0.74/3.04  &0.58/3.72  &\textbf{0.91/2.54}\\         % \\ 表示重新开始一行
2nd row     &1/3.17         &0.23/4.43          &0.90/3.61  &0.42/4.69  &0.68/4.23  &0.39/3.94  &\textbf{0.94/3.54}\\        % & 表示列的分隔线
3rd row     &1/3.46         &0.26/3.73          &0.81/3.34  &0.47/3.75  &0.62/3.89  &0.48/3.57  &\textbf{0.91/3.33}\\        % & 表示列的分隔线
2,000 test images &1/3.67   &0.74/3.53          &0.90/3.50  &0.69/4.16  &0.84/3.89  &0.63/3.66  &\textbf{0.92/3.46}\\ \hline
\end{tabular}
\end{table*}
%------------------------------------------------------------------------------
\section{Experiments}\label{section:experiments}
In this section, we elaborately construct an image dataset and spend about 10 hours on training the end-to-end network by using the Caffe software package~\cite{jia2014caffe}. To evaluate the performance of our method, we use both the synthetic test data, the public real-world dataset and compare with four recent state-of-the-art low-light image enhancement methods. At the same time, we analyse the running time and evaluate the effect of hyper-parameters to the final results.
\begin{figure*}[!t]
\centering
\subfigure[Origional image] {\includegraphics[width=0.16\linewidth]{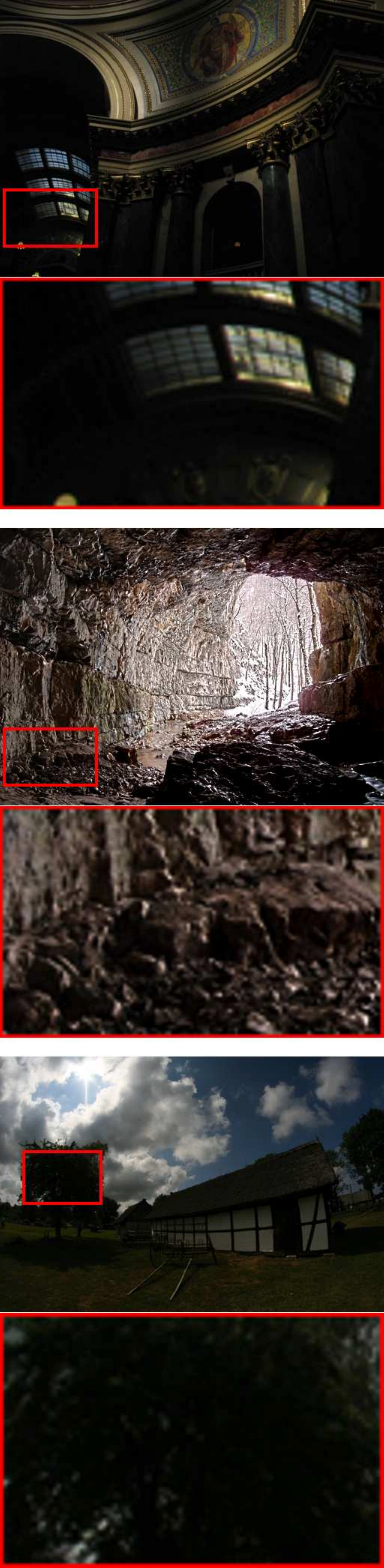}}
\subfigure[MSRCR\cite{jobson1997multiscale}] {\includegraphics[width=0.16\linewidth]{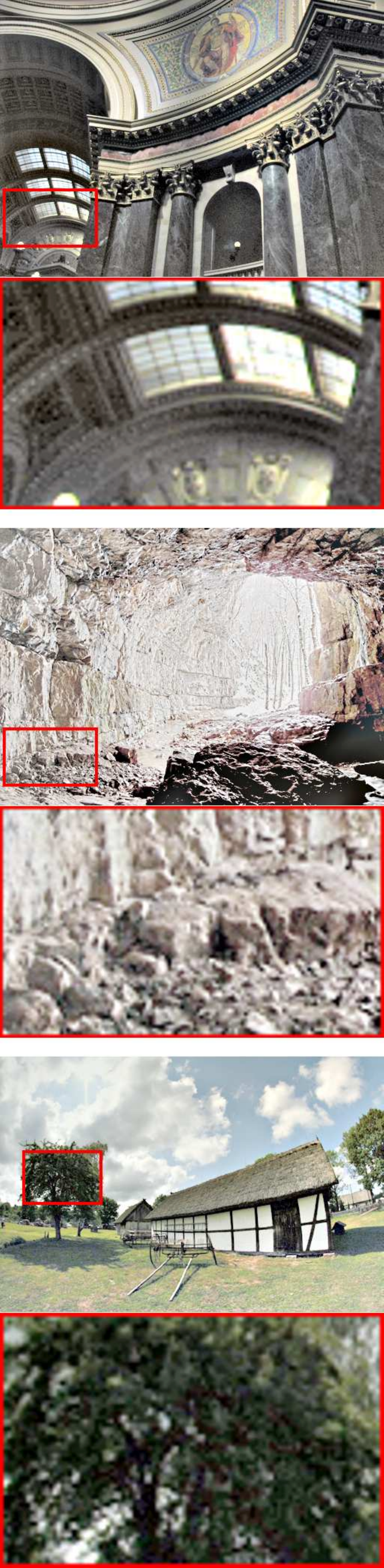}}
\subfigure[Dong\cite{dong2010fast}] {\includegraphics[width=0.16\linewidth]{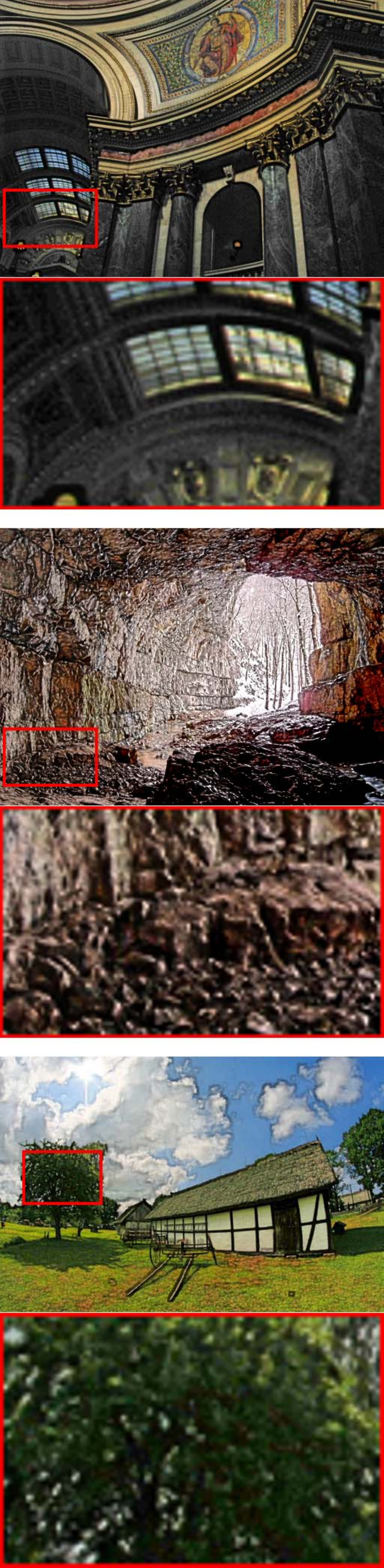}}
\subfigure[LIME\cite{guo2017lime}] {\includegraphics[width=0.16\linewidth]{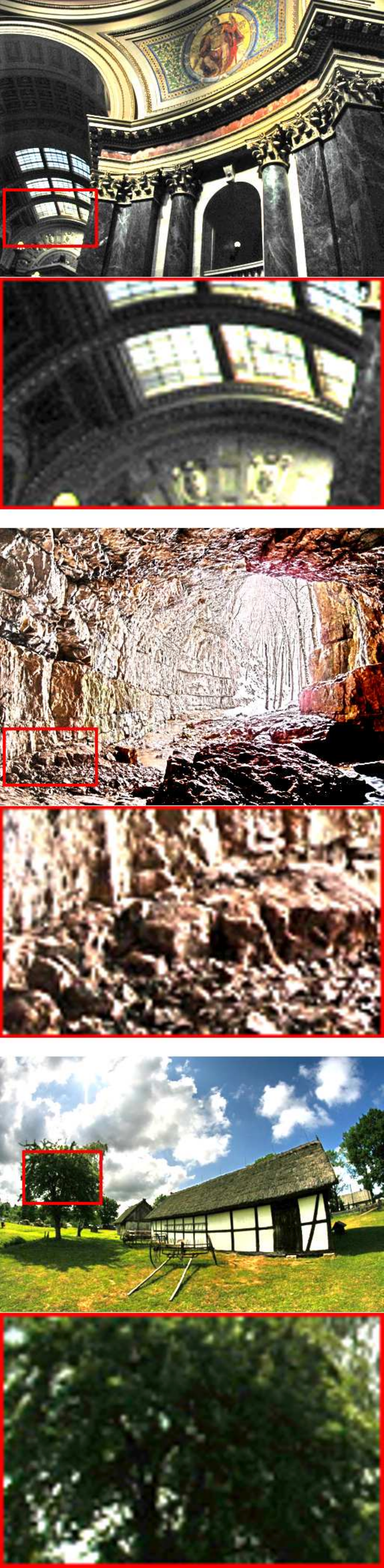}}
\subfigure[SRIE\cite{fu2016weighted}] {\includegraphics[width=0.16\linewidth]{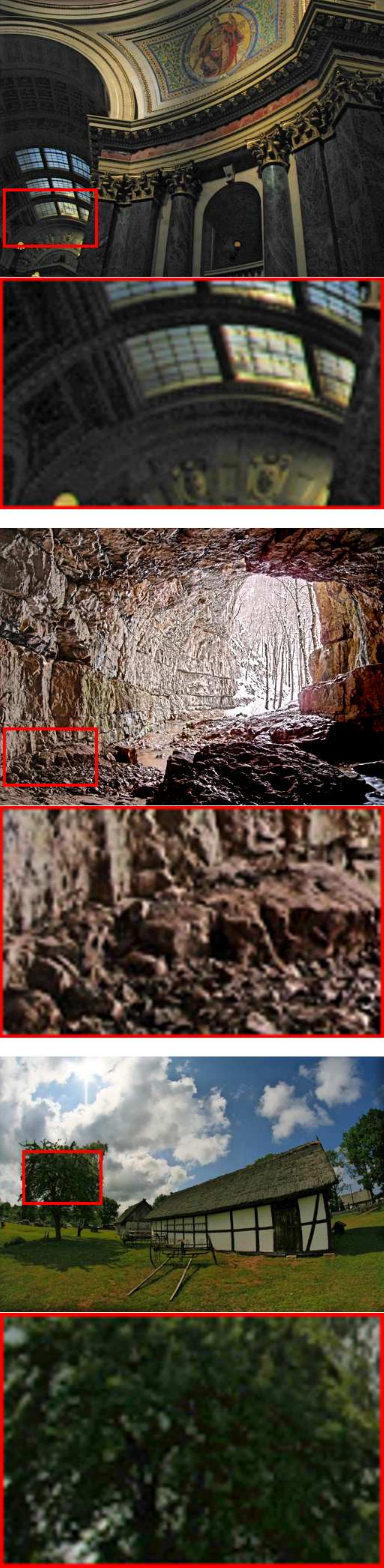}}
\subfigure[Ours] {\includegraphics[width=0.16\linewidth]{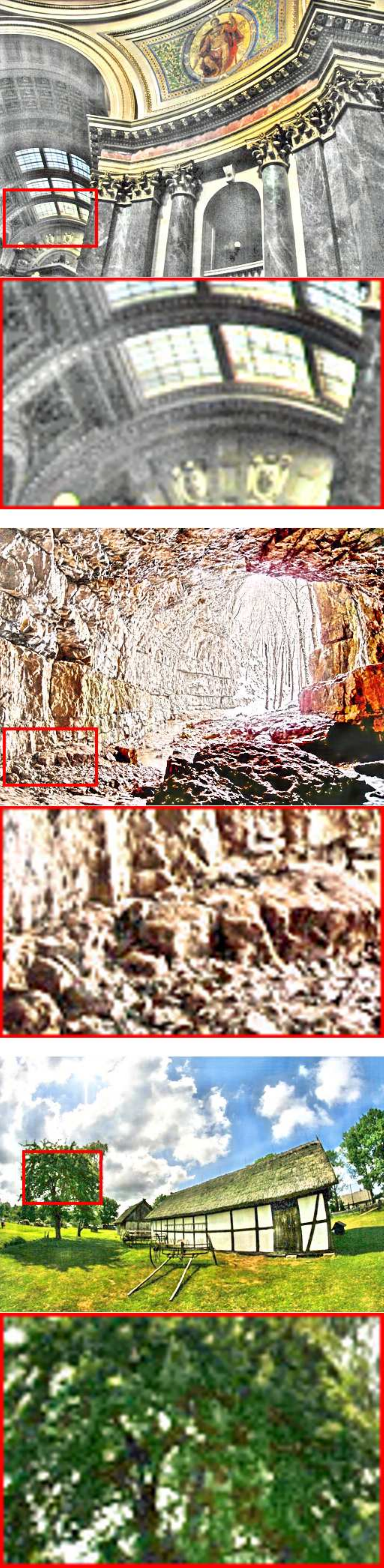}}
\caption{The results of real-world images using different methods and zoomed in region: (top-to-bottom): ``Madison”, ``Cave”, ``Kluki”.}
\label{fig:real-world images results}
\end{figure*}

\begin{table*}[tbp]
\centering  % 表居中
\caption{Quantitative measurement results using Discrete Entropy/NIQE on the real-world images}\label{tab:real-world images}
\begin{tabular}{l|ccccccc}  % {lccc} 表示各列元素对齐方式，left-l,right-r,center-c
\hline
                    &Real-world images  & MSRCR\cite{jobson1997multiscale}        &Dong\cite{dong2010fast}     & LIME\cite{guo2017lime}    & SRIE\cite{fu2016weighted}    & Ours \\ \hline  % \hline 在此
Madison            &9.56/3.05         &15.01/3.13      &11.19/3.63  &10.74/3.43  &10.97/4.18  &\textbf{15.49/3.06}\\
Cave               &13.75/3.21         &14.91/5.91      &14.54/6.35  &12.66/6.19  &14.66/\textbf{4.44}  &\textbf{15.75}/5.84\\
Kluki              &10.36/3.19         &14.62/2.33      &12.71/2.43  &12.86/2.78  &12.60/2.42  &\textbf{15.91/1.81}\\
MEF dataset       &11.02/4.31         &15.47/3.32      &13.27/4.06  &13.52/3.69  &13.16/3.63  &\textbf{16.28/3.03}\\
NPE dataset       &12.54/4.13         &14.61/4.37      &14.41/4.12  &13.93/4.25  &14.12/4.05  &\textbf{15.34/3.73}\\
VV dataset        &12.47/3.52         &17.87/2.54      &14.80/2.76  &15.79/2.48  &14.25/2.78  &\textbf{18.26/2.29}\\ \hline
\end{tabular}
\end{table*}

\subsection{Image Dataset Generation}
On the one hand, in order to learn the parameters of the MSR-net, we construct a new image dataset, which contains a great amount of high quality(HQ) and low-light(LL) natural images. An important consideration is that all the image should be selected in real world scenes. We collect more than 20,000 images from the UCID dataset~\cite{schaefer2003ucid}, the BSD dataset~\cite{arbelaez2011contour} and Google image search. Unfortunately, many of these images suffer significant distortions or contain inappropriate content. Images with obvious distortions such as heavy compression, strong motion blur, out of focus blur, low contrast, underexposure or overexposure and substantial sensor noise are deleted firstly. After this, we exclude inappropriate images such as too small or too large size, cartoon and computer generated content to obtain 1,000 better source images. Then, for each image, we use Photoshop method~\cite{photoshopessentials} to figure out the ideal brightness and contrast settings, then process them one by one to get the high quality(HQ) images with the best visual effect. At last, each HQ image is used to generate 10 low-light(LL) images by reducing brightness and contrast randomly and using gamma correction with stochastic parameters. So we attain a dataset containing 10,000 pairs of HQ/LL images. Further, 8,000 images in the dataset are randomly selected to generate one million $64 \times 64$ HQ/LL patch pairs for training. And the remaining 2,000 images pairs are used to test the trained network during training(please see more details about the dataset generation in the supplemental materials).

On the other hand, in order to evaluate our approach objectively and impartially, we choose the real-world low-light images form the public MEF dataset~\cite{ma2015perceptual,zeng2014perceptual}, NPE dataset~\cite{wang2013naturalness} and VV dataset~\cite{vonikakis2012improving,vonikakis2013biologically}.
\subsection{Training Setup}
We set the depth of MSR-net to $K=10$, and use Adam with weight decay of $10^{-6}$ and a mini-batch size of 64. We start with a learning rate of $10^{-4}$, dividing it by 10 at 100K and 200K iterations, and terminate training at 300K iterations. During our experiments, we found that the network with multi-scale logarithmic transformation performs better than that with single-scale logarithmic transformation, so we set the number of logarithmic transformation function $n=4$ and $v=1,10,100,300$ respectively. The size of convolution kernel has been described partially in the previous section~\ref{section:Proposed method}, and the specific values are shown in the Table~\ref{tab:MSR-net configuration}.
\begin{table}[htbp]
\centering
\caption{MSR-net network configuration. The convolutional layer parameters are denoted as ``conv$\langle$receptive field size$\rangle$-$\langle$numbers of output channels$\rangle$". The ReLU activation is not shown for brevity}
\label{tab:MSR-net configuration}
\begin{tabular}{|c|c|c|c|}
\hline
$W_{-1}$  &conv1-32 &$W_{11}$ &conv1-3\\         % \\ 表示重新开始一行
\hline
$W_0$  &conv3-3 &$W_{12}$ &conv1-3\\        % & 表示列的分隔线
\hline
$W_i,i=1,...,10$  &conv3-32 & & \\
\hline
\end{tabular}
\end{table}

\subsection{Results on synthetic test data}
Figure~\ref{fig:synthesized test image} shows visual comparison for three synthesized low light images. As we can see, the result of MSRCR~\cite{jobson1997multiscale} looks unnatural, the method proposed by Dong~\cite{dong2010fast} always generates unexpected black edge and the result of SRIE~\cite{fu2016weighted} tends to be dark in some extent. LIME~\cite{guo2017lime} has a similar result to our method, while ours achieves better performance in dark regions.

Since the ground truth is known for the synthetic test data, we use SSIM~\cite{wang2004image} for a quantitative evaluation and NIQE~\cite{mittal2013making} to assess the natural preservation. A higher SSIM indicates that the enhanced image is closer to the ground truth, while a lower NIQE value represents a higher image quality. All the best results are boldfaced. As shown in Table~\ref{tab:synthesized test images}, our method achieves higher SSIM and lower NIQE average than other methods for 2,000 test images.

\subsection{Results on real-world data}
Figure~\ref{fig:real-world images results} also shows the visual comparison for three real-world low-light images. As shown in every red rectangle, our method MSR-net always achieves better performance in dark regions. More specifically, in the first and second image we get brighter result. In the third image we achieve more natural result, for instance, the tree has been enhanced to be bright green. Besides, from the Garden image in Figure~\ref{fig:introduction figure}, our result gets a clearer texture and richer details than other methods.

Besides the NIQE to evaluate the image quality, we assess the detail enhancement through the Discrete Entropy~\cite{ye2007discrete}. A higher discrete entropy shows that the color is richer and the outline is clearer. We delete the high-light images and only keep the low-light images on the MEF dataset~\cite{ma2015perceptual,zeng2014perceptual}, NPE dataset~\cite{wang2013naturalness} and VV dataset~\cite{vonikakis2012improving,vonikakis2013biologically} to evaluate our method. As shown in Table~\ref{tab:real-world images}, for different dataset, MSR-net can also obtain lower NIQE and higher discrete entropy.

\begin{figure}[h]
  \centering
\subfigure[Real-world image] {\includegraphics[width=0.32\linewidth]{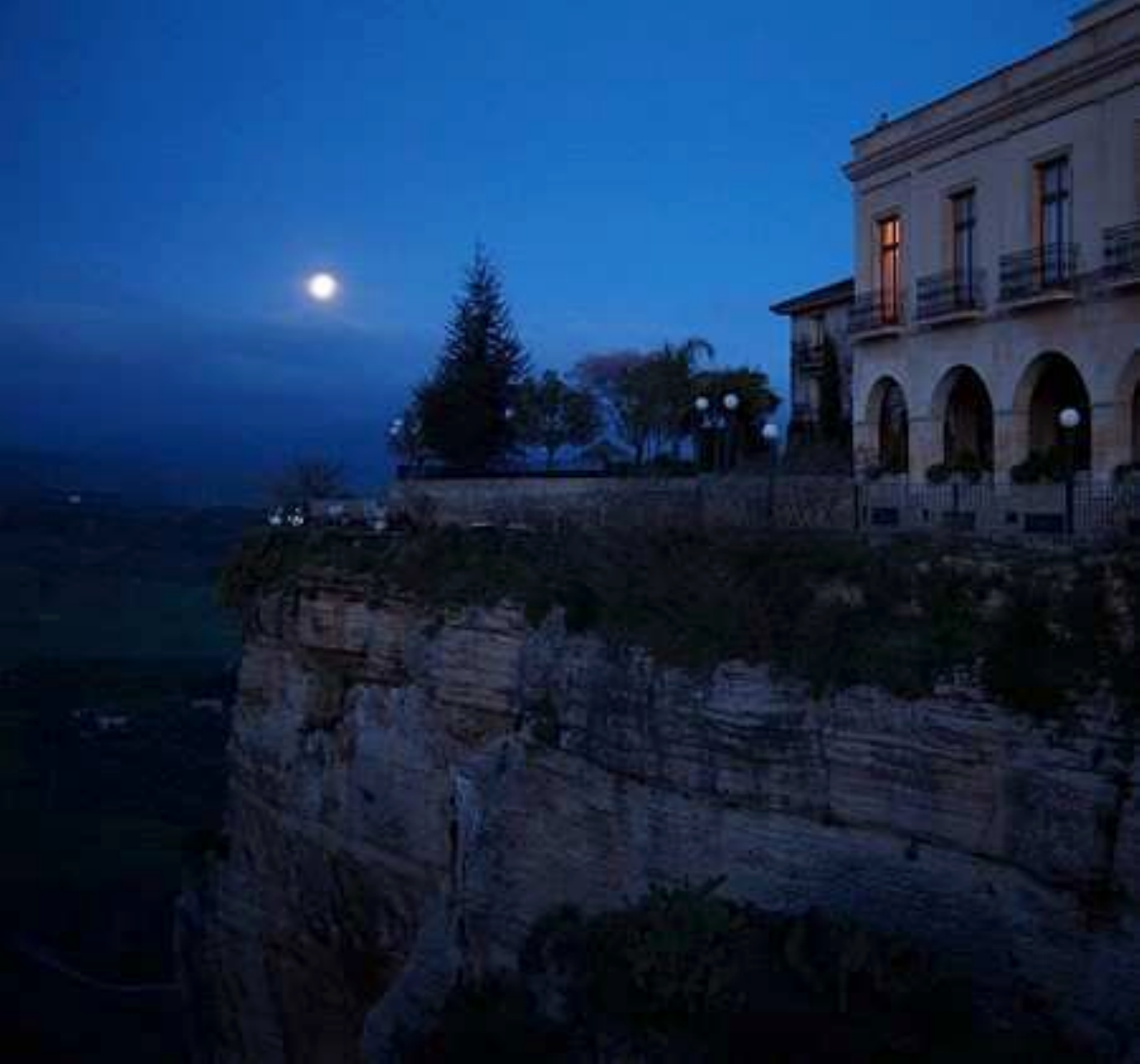}}
\subfigure[Without denoising] {\includegraphics[width=0.32\linewidth]{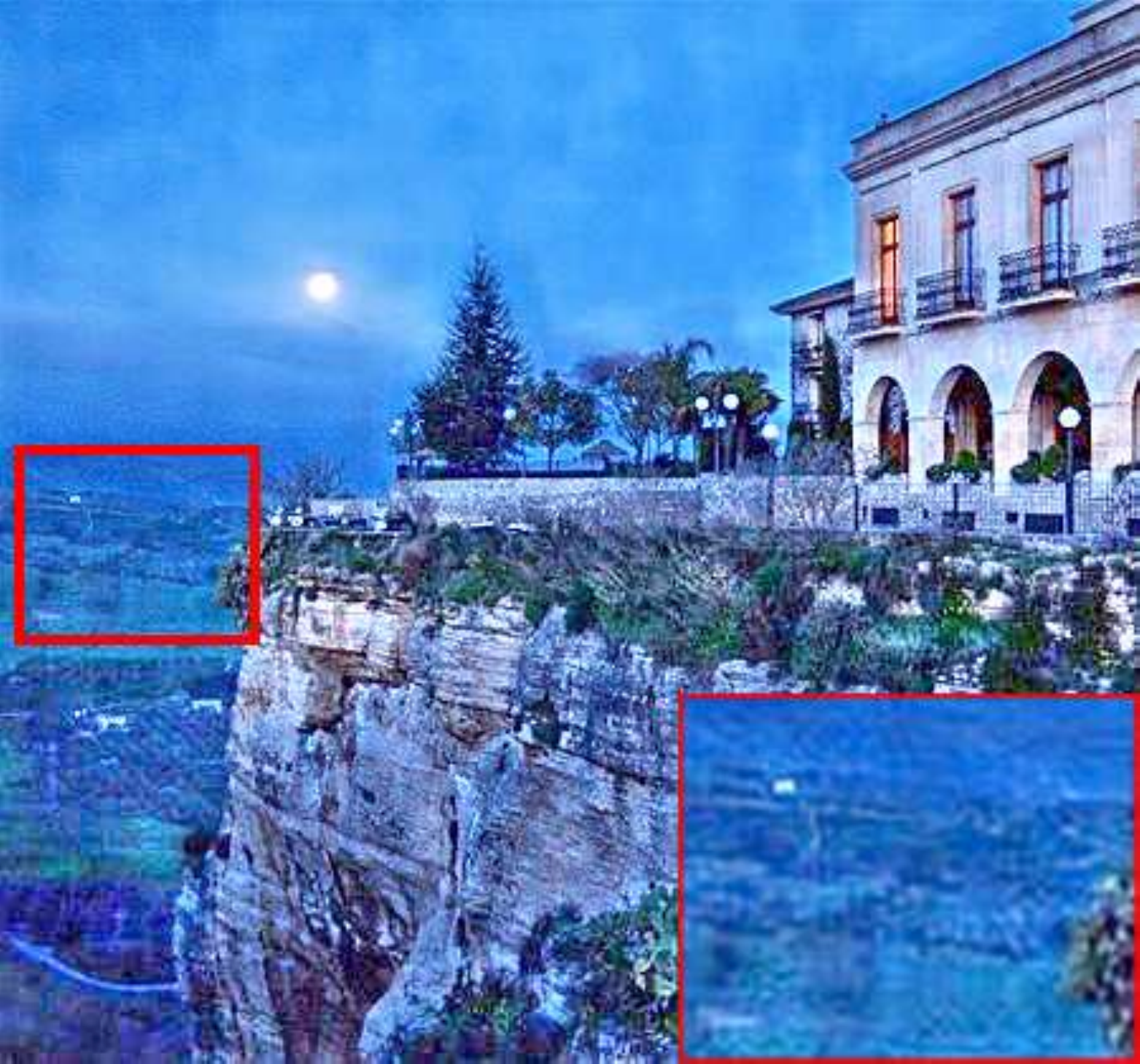}}
\subfigure[With denoising] {\includegraphics[width=0.32\linewidth]{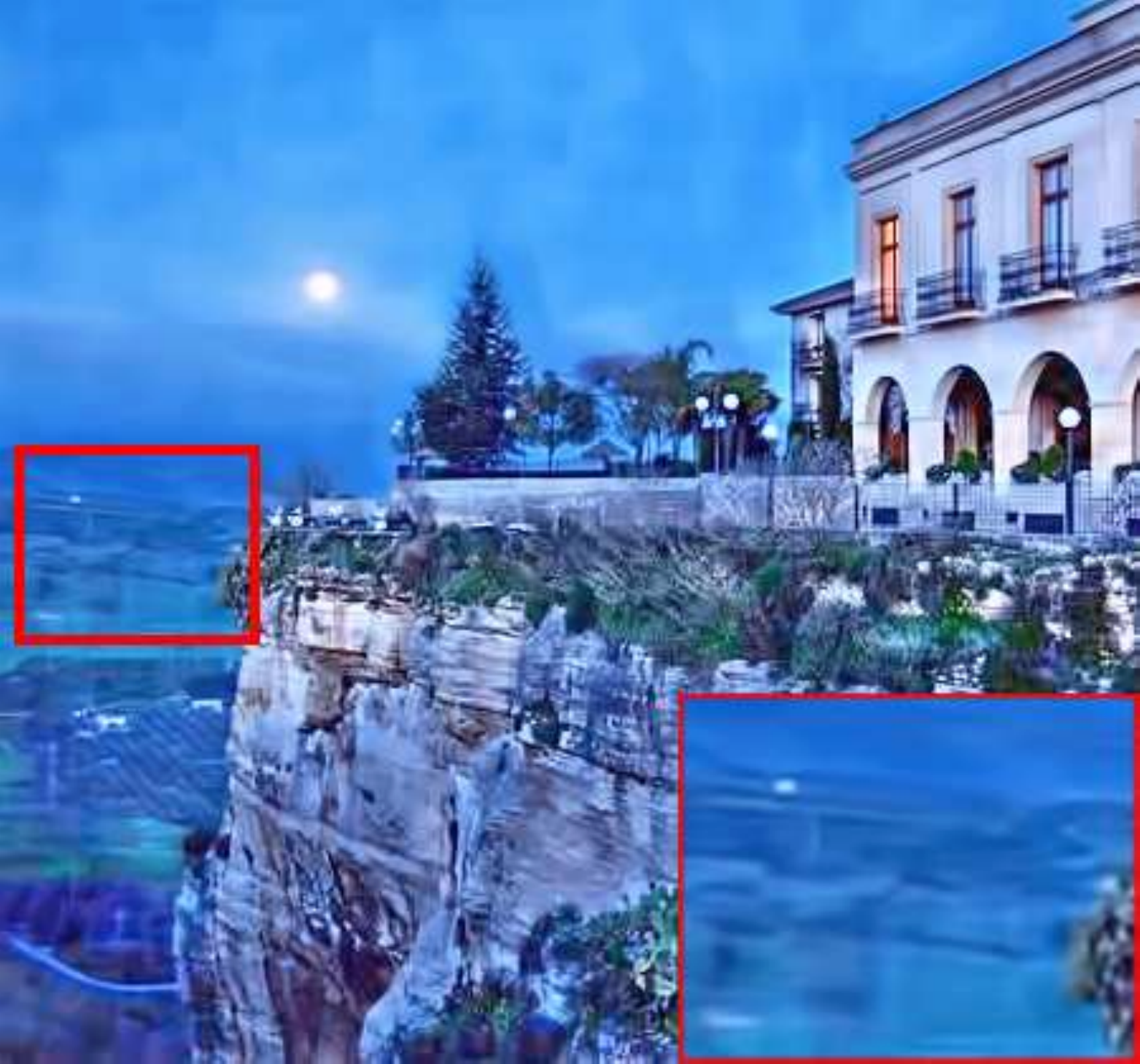}}
  \caption{Results without and with denoising}
  \label{fig:denosing}
\end{figure}

Considering the fact that dealing with real-world low light images sometimes causes noise, we attempt to use a denoising algorithm BM3D~\cite{dabov2008image} as a post-processing. An example is shown in Figure~\ref{fig:denosing}, where removing the noise after our deep network can further improve the visual quality on real-world low light image.

\subsection{Color Constancy}
In addition to enhance the dark image, our model also does a good job in correcting the color. Figure~\ref{fig:synthesized test image} provides some examples. As we can see, our enhanced image is much more similar to the ground truth. To evaluate the performance of the different algorithms, the angular error~\cite{Hordley2004Re} between the ground truth image $Y$ and model result $\hat Y$ is used:
\begin{equation}
\varepsilon  = \arccos \left( {\frac{{ < Y,\hat Y > }}{{||Y|| \cdot ||\hat Y||}}} \right)
\end{equation}

A lower error indicates that the color of enhanced image is closer to the ground truth. Table~\ref{tab:color constancy} gives some specific result of the images in Figure~\ref{fig:synthesized test image}.
\begin{table}[htbp]
\centering
\caption{Angular error $\varepsilon$ (degree) of different model}\label{tab:color constancy}
\begin{tabular}{l|ccccc}
\hline
                    & Dong\cite{dong2010fast}     & LIME\cite{guo2017lime}    & SRIE\cite{fu2016weighted}    & Ours \\ \hline  % \hline 在此
1st row             &18.67           &16.49  &20.01  &\textbf{5.89}\\         % \\ 表示重新开始一行
2nd row              &33.62          &25.40  &35.51  &\textbf{6.55}\\        % & 表示列的分隔线
3rd row               &25.77          &21.55  &24.73  &\textbf{6.20}\\
2K images          &18.45          &13.53  &17.59  &\textbf{5.79}\\     \hline      % & 表示列的分隔线
\end{tabular}
\end{table}

\begin{table*}[htbp]
\centering
\caption{Comparison of average running time on 100 images(seconds)}\label{tab:comparison of average running time}
\begin{tabular}{c|cccccc}
\hline
Image size          & MSRCR\cite{jobson1997multiscale}       &Dong\cite{dong2010fast}   & LIME\cite{guo2017lime}    & SRIE\cite{fu2016weighted}     & Ours(CPU)    &Ours(GPU)\\ \hline
$500\times500$      &0.761                                   &0.321                     &\textbf{0.188}             &1.370                           &2.422        &0.320  \\
$750\times750$      &1.266                                   &0.681                     &\textbf{0.359}             &3.090                           &5.043        &0.678   \\
$1,000\times1,000$  &1.668                                   &1.103                     &\textbf{0.521}             &5.755                         &8.962             &0.824  \\ \hline
\end{tabular}
\end{table*}
\subsection{Running time on test data}
Compared with other non-deep methods, our approach processes the low-light images efficiently. Table~\ref{tab:comparison of average running time} shows the average running time of processing a test image for three different sizes, and each averaged 100 testing images. These experiments are tested on a PC running Windows 10 OS with 64G RAM, 3.6GHz CPU and Nvidia GeForce GTX 1080 GPU. All codes of these methods are run in Matlab, which ensures the fairness of time comparison. Methods~\cite{jobson1997multiscale},\cite{dong2010fast},\cite{guo2017lime},\cite{fu2016weighted} are implemented using CPU, while our method is tested on both CPU and GPU.
Because our method is a completely feedforward process after network training, we can find that our approach on GPU processes significantly faster than methods~\cite{jobson1997multiscale},\cite{dong2010fast},\cite{fu2016weighted} except \cite{guo2017lime}.

\subsection{Study of MSR-net Parameters}
The number of logarithmic transformation function $n$ and the number of convolutional layers $K$ are two main hyper-parameters in MSR-net. In this subsection, we try to experiment on the effect of these hyper-parameters on the final results. As we all know, the effectiveness of deeper structures for low-level image tasks is found not as apparent as that shown in high-level tasks~\cite{Bruna2015Image,Ren2016Single}. Specifically, we test for the number of logarithmic transformation function $n \in \left\{ {1,2,4} \right\}$ and $v=300$;$v=100,300$;$v=1,10,100,300$ respectively. At the same time, we set the number of convolutional layers $K \in \left\{ {6,10,21} \right\}$. For the sake of fairness, all these networks are iterated 100K times and 100 synthetic images are used to measure the result by averaging their SSIM.

\begin{figure}[h]
  \centering
\subfigure[Ground truth / SSIM] {\includegraphics[width=0.45\linewidth]{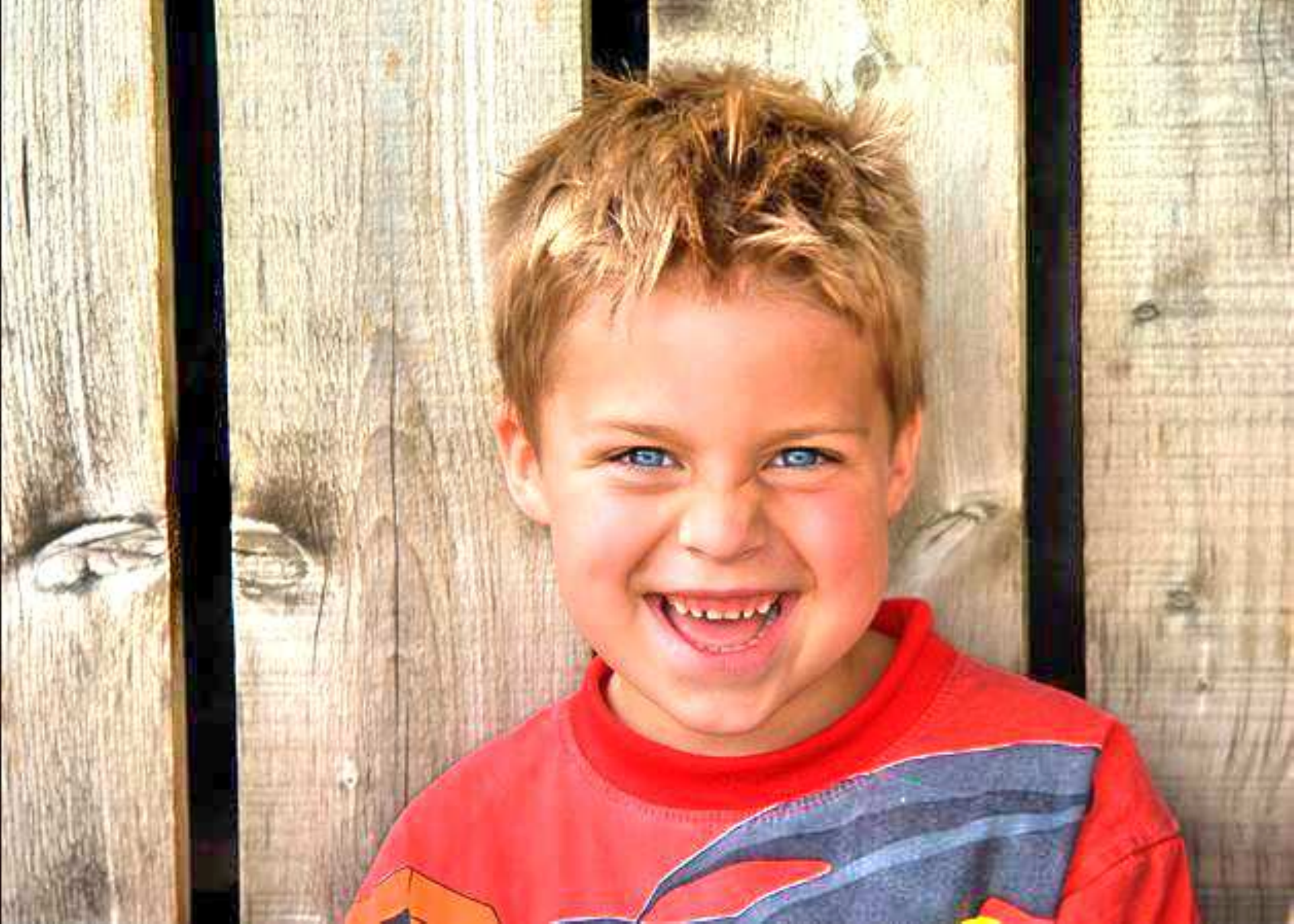}}
\subfigure[Synthetic test data / 0.29] {\includegraphics[width=0.45\linewidth]{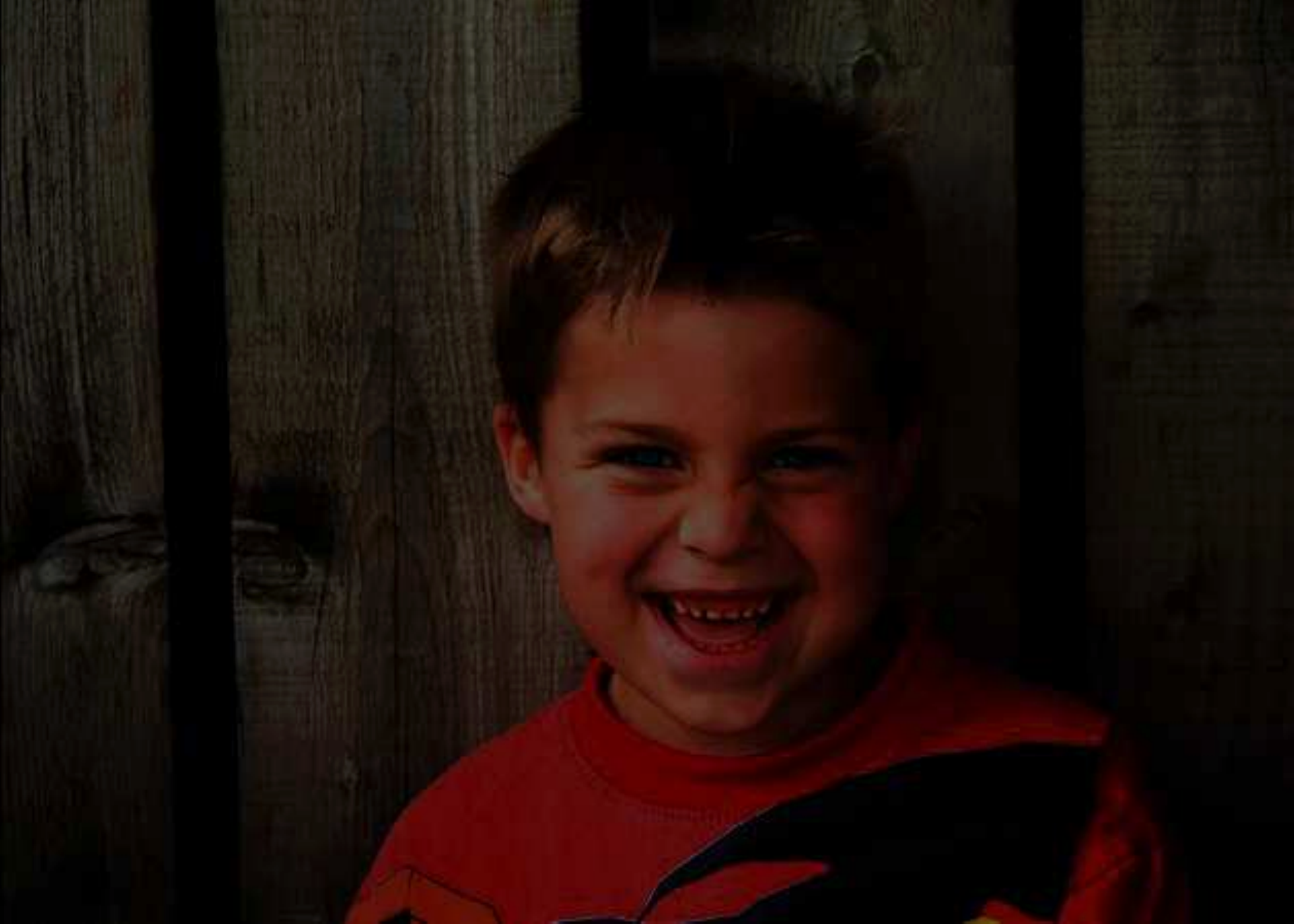}}
\subfigure[$n=1,K=10$ / 0.84] {\includegraphics[width=0.45\linewidth]{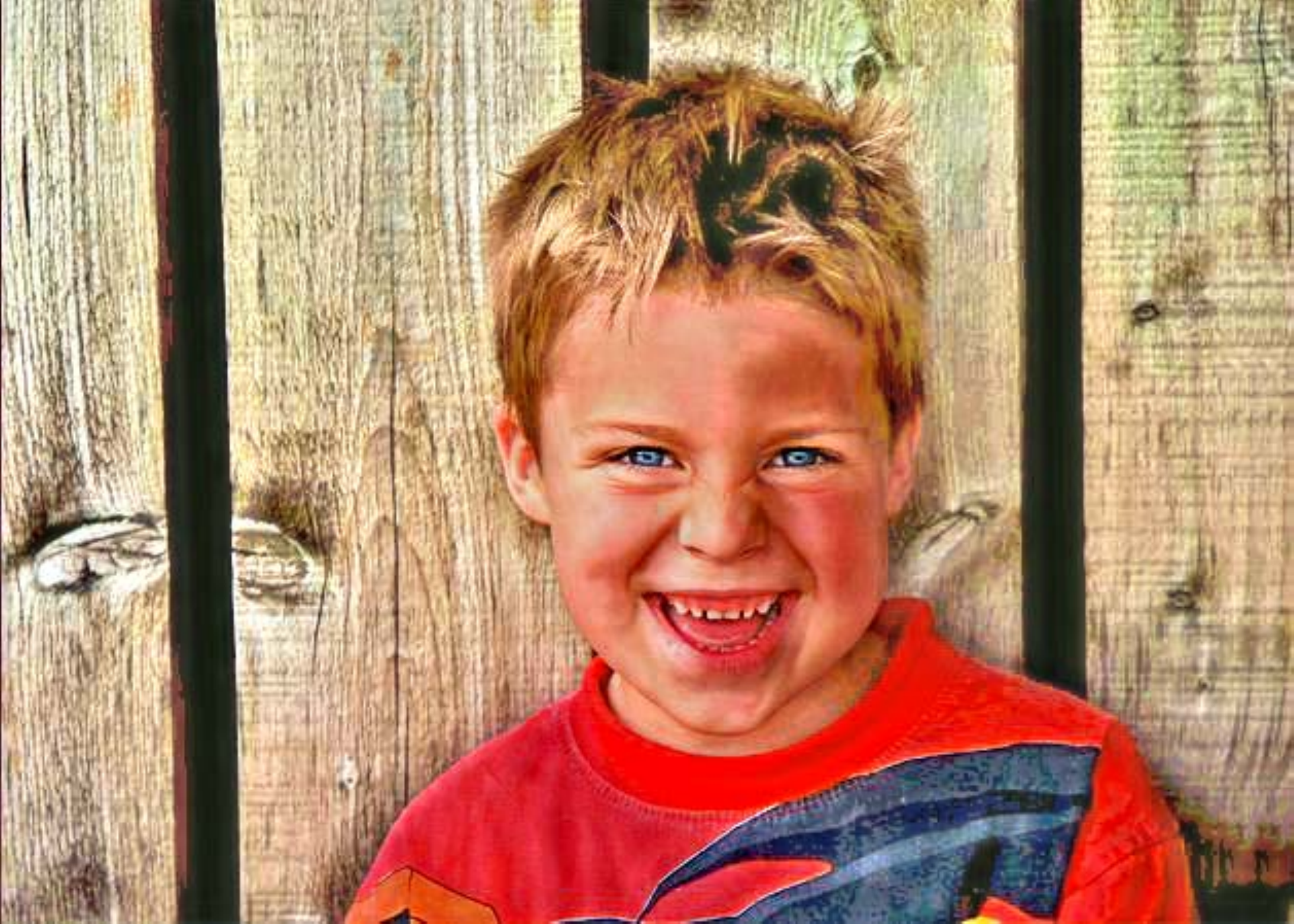}}
\subfigure[$n=4,K=10$ / 0.89] {\includegraphics[width=0.45\linewidth]{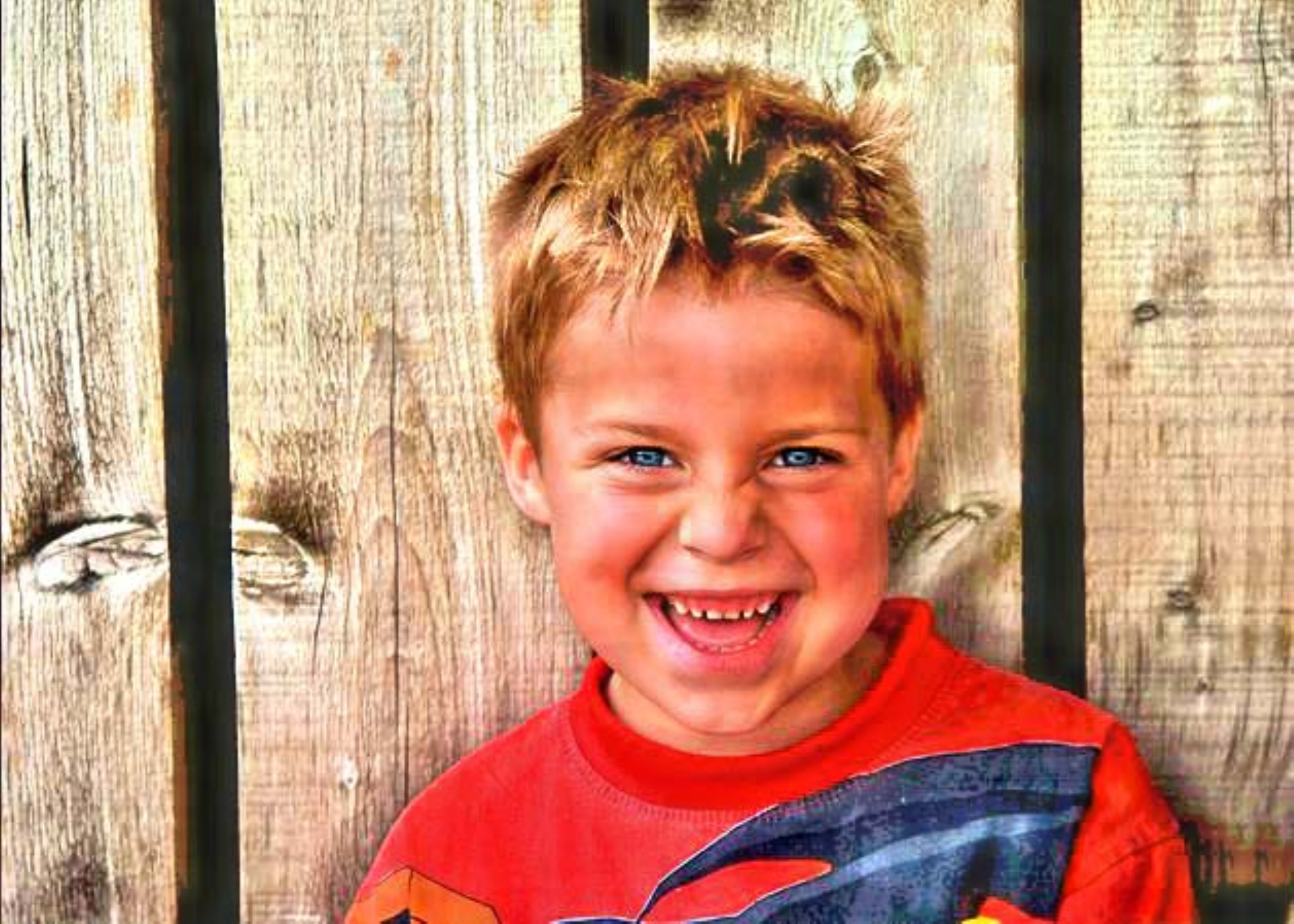}}
  \caption{Results using different parameter setting}
  \label{fig:Comparison with N}
\end{figure}

As shown in Table~\ref{Tab:average ssim}, adding more hidden layers obtains higher SSIM and achieves better results. We believe that, with an appropriate design to avoid over-fitting, deeper structure can improve the network's nonlinear capacity and learning ability. In Figure~\ref{fig:Comparison with N}, from the color of the boy's skin and the clothes, the network using multi-scale logarithmic transformation performs better. It is also essential for MSR-net to improve nonlinear capacity by using multi-scale logarithmic transformation.
To get better performance within the running time and hardware limits, we finally chose the number of logarithmic transformation function $n=4$ and the depth of convolutional layers $K=10$ for our experiments above.

\begin{table}[htbp]
\centering
\caption{Average SSIM for different network parameters}
\label{Tab:average ssim}
\begin{tabular}{l|ccc}
\hline
 &$K=6$ &$K=10$ &$K=21$\\ \hline  % \hline 在此行下面画一横线
$n=1$  &0.879 &0.897 &0.904\\         % \\ 表示重新开始一行
$n=2$  &0.890 &0.902 &0.914\\        % & 表示列的分隔线
$n=4$  &0.893 &0.905 &0.920\\ \hline
\end{tabular}
\end{table}

\section{Conclusion}
In this paper, we propose a novel deep learning approach for low-light image enhancement. It shows that multi-scale Retinex is equivalent to a feedforward convolutional neural network with different Gaussian convolution kernels. After this, we construct a Convolutional Neural Network(MSR-net) that directly learns an end-to-end mapping between dark and bright images with little extra pre/post-processing beyond the optimization. Experiments on synthetic and real-world data reveal the advantages of our method in comparison with other state-of-the-art methods from the qualitative and quantitative perspective. Nevertheless, there are still some problems with this approach. Because of the limited receptive field in our model, very smooth regions such as clear sky are sometimes attacked by halo effect. Enlarging receptive field or adding hidden layers may solve this problem.

{\small
\bibliographystyle{ieee}
\bibliography{IECNNbib}
}

\end{document}